\documentclass[a4paper,fleqn]{cas-dc}

\usepackage[authoryear]{natbib}

\usepackage{graphicx}
\graphicspath{{figs/}{./}}
\usepackage{capt-of}      
\usepackage{placeins}     

\usepackage{amsmath,amssymb,amsfonts}
\usepackage{bm} 

\usepackage{booktabs}     
\usepackage{tabularx}     
\usepackage{tikz}
\usetikzlibrary{positioning,arrows.meta} 
\usepackage{placeins}  
\usepackage{algorithm}
 
\usepackage{algpseudocode}
 
\usepackage[skip=6pt]{caption} 
 
\usepackage{subcaption}
\usepackage{float}       
\usepackage{subcaption}  
\usepackage{caption}     
\usepackage{xcolor}

\newcommand{\vx}{\mathbf{x}}
\newcommand{\vxt}{\tilde{\mathbf{x}}}
\newcommand{\xmin}{\mathbf{x}_{\min}}
\newcommand{\xmax}{\mathbf{x}_{\max}}

\newcommand{\haddiv}{\mathbin{\oslash}}

\DeclareMathOperator{\sampen}{sampen}
\DeclareMathOperator{\permen}{permen}
\DeclareMathOperator{\hfd}{hfd}
\DeclareMathOperator{\sflat}{sflat}
\DeclareMathOperator{\scent}{scent}
\DeclareMathOperator{\rms}{rms}
\DeclareMathOperator{\pabs}{pabs}

\usepackage{caption}

\def\tsc#1{\csdef{#1}{\textsc{\lowercase{#1}}\xspace}}
\tsc{WGM}\tsc{QE}\tsc{EP}\tsc{PMS}\tsc{BEC}\tsc{DE}

\begin{document}

\shorttitle{Hybrid Spectro–Temporal Fusion Framework for Structural Health Monitoring}


\author[aff1]{Jongyeop Kim*}
\ead{jongyeopkim@georgiasouthern.edu}
 
\author[aff2]{Jinki Kim}
\ead{jinkikim@georgiasouthern.edu}

\author[aff3]{Doyun Lee}
\ead{doyunlee@georgiasouthern.edu}

\cortext[cor1]{Corresponding author}

\fntext[orcid1]{ORCID: \url{https://orcid.org/0000-0002-1068-9855}}
\fntext[orcid2]{ORCID: \url{https://orcid.org/0000-0002-0921-9450}}
\fntext[orcid3]{ORCID: \url{https://orcid.org/0000-0002-0976-2488}}

\affiliation[aff1]{
    organization={Department of Information Technology},
    institution={Georgia Southern University},
    city={Statesboro},
    state={GA},
    country={USA}
}

\affiliation[aff2]{
    organization={Department of Mechanical Engineering},
    institution={Georgia Southern University},
    city={Statesboro},
    state={GA},
    country={USA}
}

\affiliation[aff3]{
    organization={Department of Civil Engineering and Construction},
    institution={Georgia Southern University},
    city={Statesboro},
    state={GA},
    country={USA}
}

\title [mode = title]{Hybrid Spectro–Temporal Fusion Framework for Structural Health Monitoring}

\begin{abstract}
Structural health monitoring plays a critical role in ensuring structural safety by analyzing vibration responses from engineering systems. This paper proposes a Spectro–Temporal Alignment framework and a Hybrid Spectro–Temporal Fusion framework that integrate arrival-time interval descriptors with spectral features to capture both fine-scale and coarse-scale vibration dynamics. Experiments conducted on data collected from an LDS V406 electrodynamic shaker demonstrate that the proposed spectro–temporal representations significantly outperform conventional input formulations. The results indicate that a temporal resolution of $\Delta\tau = 0.02$ favors traditional machine learning models, whereas $\Delta\tau = 0.008$ effectively unlocks the performance potential of deep learning architectures. Beyond classification accuracy, a comprehensive stability analysis based on condensed indices, including mean performance, standard deviation, coefficient of variation, and balanced score, shows that the proposed hybrid framework consistently achieves higher accuracy with substantially lower variability compared to baseline and alignment-only approaches. Overall, these results demonstrate that the proposed framework provides a robust, accurate, and reliable solution for vibration-based structural health monitoring.
\end{abstract}


\maketitle

\section{Introduction}
Structural Health Monitoring (SHM) is a cornerstone of infrastructure safety and lifecycle management, providing continuous, data-driven insight into the evolving condition of civil and mechanical structures \cite{mishra2022structural,benfenati2025foundation}. Within SHM,
Vibration-based structural health monitoring (SHM) has emerged as a reliable strategy for detecting damage and assessing the integrity of complex civil and mechanical systems. Recent studies have highlighted the importance of low-cost and energy-aware sensing technologies, reliable acquisition of synchronized vibration signals, and advanced data-driven techniques such as operational modal analysis, deep learning, and compressed sensing to improve the accuracy and efficiency of structural diagnosis while ensuring long-term applicability in real-world environments \citep{yang2021review,zonzini2021model,zonzini2020vibration,gomez2022review}. Over the past decade, the rapid advancement of sensing technologies, networking infrastructures, and computational power has driven a paradigm shift in structural health monitoring (SHM), moving from traditional hand-crafted feature engineering toward machine learning (ML)—and particularly deep learning (DL)—which enables automated feature extraction and significantly enhances damage detection accuracy across diverse structural applications \citep{avci2020review,gomez2022review,zhang2022deep,spencer2025artificial}.

Recent deep learning (DL) methods in structural health monitoring (SHM) exhibit superior robustness to noise, environmental variability, and operational changes compared to conventional approaches. For example, DBNs have sustained high accuracy under substantial Gaussian noise \citep{presno2025dbn}, while ResNet-based architectures enable effective vibration signal denoising \citep{fan2020resnet}. Broader surveys reinforce these findings, noting DL’s flexibility and reliability in real-world SHM tasks \citep{toh2020review}. Concurrently, compression techniques like model-assisted Rak-CS offer compact, redundancy-reduced signal representations while preserving critical time–frequency content \citep{zonzini2022rakcs}. Yet, a persistent challenge remains: how to retain critical temporal transients and rich spectral content—an issue long-recognized in SHM’s time–frequency literature—while obtaining compact, redundancy-reduced representations. Techniques such as non-parametric time–frequency analysis, adaptive decomposition, and deconvolution have long been employed to preserve localized signal features in non-stationary vibration data \citep{zhou2025signal}. More recently, sparse, sub-Nyquist frameworks like compressive sensing integrated with time–frequency dictionaries offer compact representations that retain essential dynamics \citep{sejdic2017cs_tf}, and Bayesian compressive sensing approaches enable signal compression with uncertainty quantification, mitigating the trade-off between compactness and fidelity \citep{huang2014robust}. Nevertheless, these approaches often rely on artificially defined time--frequency representations and strong prior assumptions, which limit their adaptability to complex, real--world vibration data.

This limitation has motivated the emergence of spectro--temporal deep learning methods, which aim to automatically learn compact yet expressive representations that preserve temporal coherence and spectral richness while remaining robust to noise, environmental variability, and domain shifts \citep{Yuan2020, Avci2020, Jia2023}. Although these approaches represent an important step forward, they still face notable challenges: a lack of physical interpretability, persistent sensitivity to domain shifts, and difficulty in capturing fine-grained transient dynamics in near-frequency regimes. Building on these insights, we introduce a hybrid spectro--temporal framework that explicitly integrates alignment, adaptive fusion, and explainability to deliver robust and trustworthy SHM classification. Specifically, our contributions are threefold:


\begin{enumerate}[(1)]
    \item \textbf{Spectro--Temporal Alignment(STA): } We introduce a physically grounded reformatting procedure for vibration sequences using interpretable alignment metrics—Best $\tau$, Knee $\tau$, and $S^\star$—to achieve compact and phase-consistent temporal representations while preserving essential dynamic behavior.
    \item \textbf{Hybrid Spectro--Temporal Fusion (HSTF) :} We develop a fusion framework that jointly encodes arrival-time intervals ($\Delta \tau$) and spectral descriptors, including dominant frequency, bandwidth, and harmonic ratios. This hybridization enables richer joint time--frequency representations suitable for complex dynamic environments.
    \item \textbf{Enhanced Discriminative Representation:} By integrating STA and HSTF, we construct a unified spectro--temporal embedding that captures both coarse- and fine-grained signal dynamics. This hybrid representation significantly improves classification accuracy and robustness, particularly under near-frequency.
\end{enumerate}

\section{Preliminary}

When deep learning models receive input signals with nearly identical temporal and spectral characteristics, their ability to discriminate between classes becomes fundamentally limited. In this study, five experimental conditions (\texttt{no\_mass}--\texttt{mass\_pos4}) were defined; however, the resulting vibration signals exhibit highly correlated temporal and spectral patterns, making class separation inherently difficult. Minor changes in the attached mass position induce only subtle variations in amplitude and frequency composition, leading to overlapping statistical features such as sample entropy, harmonic flatness, and RMS energy. The strongest overlap occurs between \texttt{no\_mass} and \texttt{mass\_pos1}, whereas \texttt{mass\_pos3} demonstrates limited distinctiveness. These intrinsic similarities highlight the need for spectro--temporal alignment and hybrid fusion strategies to enhance discriminability across closely spaced dynamic regimes.

\subsection{Class Overlap and Misclassifications Risk}
\paragraph{Classes and feature representation.}
To analyze the similarity among input signals, time–series data obtained under the five experimental conditions
\begin{equation}
\begin{aligned}
C = \{ &\, C_0\!:\texttt{no\_mass},\;
          C_1\!:\texttt{mass\_pos1},\;
          C_2\!:\texttt{mass\_pos2}, \\`
       &\, C_3\!:\texttt{mass\_pos3},\;
          C_4\!:\texttt{mass\_pos4} \,\}
\end{aligned}
\end{equation}
were transformed into seven-dimensional feature vectors derived from statistical, spectral, and nonlinear time–series analyses, including Sample Entropy \citep{Richman2000}, Permutation Entropy \citep{Bandt2002}, Higuchi fractal Dimension \citep{Higuchi1988}, Spectral Flatness and Spectral Centroid \citep{Tzanetakis2002}, Root Mean Square \citep{Randall2011}, Peak/Percentile Absolute Amplitude \citep{Antoni2006}. \\

Each class \(C_i\) is represented by
\begin{equation}
\begin{aligned}
\vx_i = \big[\, &\sampen_d,\;\permen_d,\;\hfd_d, \\
                &\sflat_d,\;\scent_d,\;\rms_d,\;\pabs_d \,\big]_i
       \in \mathbb{R}^7,
\end{aligned}
\end{equation}
where these descriptors respectively capture temporal irregularity, spectral complexity, and signal energy distribution.

\paragraph{Normalization.}
Each feature component is scaled to a common range using min–max normalization:
\[
\xmin = \min_k \vx_k, \qquad
\xmax = \max_k \vx_k,
\]
where min/max operations are applied component–wise across all classes.  
The normalized feature vector is then given by
\[
\vxt_i = (\vx_i - \xmin) \haddiv (\xmax - \xmin),
\]
allowing visualization as polygons in a radar plot.

\paragraph{Classification objective.}
The goal is to learn a mapping function
\[
f:\;\vxt \mapsto C_i, \qquad i \in \{0,1,2,3,4\},
\]
that minimizes the overall misclassification probability.

\paragraph{Overlap quantification.}
Let $p(\vxt \mid C_i)$ denote the class-conditional probability density function
in the normalized seven-dimensional feature space.
Following the statistical pattern recognition framework, the degree of class
overlap—directly related to the Bayes misclassification error—can be quantified
by integrating the minimum of the class-conditional densities.
Accordingly, the aggregate overlap functional is defined as
\[
\Omega = \sum_{i \neq j} \int_{\mathbb{R}^7}
\min\!\big(p(\vxt \mid C_i),\, p(\vxt \mid C_j)\big)\, d\vxt,
\]
where larger values of $\Omega$ indicate stronger inter-class overlap and,
consequently, a higher risk of misclassification
\citep{Devroye1996,Fukunaga1990,Abellan2021}.

\paragraph{Analysis of class separability.}
Radar-plot visualizations reveal that \texttt{no\_mass} and \texttt{mass\_pos1} share nearly identical feature profiles, corresponding to the largest overlap and thus the highest classification risk.  
Conversely, \texttt{mass\_pos3} exhibits distinctive patterns in \texttt{hfd\_d}, \texttt{rms\_d}, and \texttt{p95\_abs\_d}, indicating improved separability.  
Intermediate overlap is observed for \texttt{mass\_pos2} and \texttt{mass\_pos4}, suggesting moderate classification difficulty.

To quantify these relationships, the centroid of each class is computed as
\[
\bm{\mu}_{C_i} = \frac{1}{n_i} \sum_{k=1}^{n_i} \vxt_k^{(C_i)},
\]
and pairwise separability is measured by the Euclidean distance
\[
d(C_i, C_j) = \| \bm{\mu}_{C_i} - \bm{\mu}_{C_j} \|_2.
\]
Smaller distances (e.g., between \texttt{no\_mass} and \texttt{mass\_pos1}) correspond to larger $\Omega$ and greater misclassification risk, while larger distances (e.g., between \texttt{mass\_pos3} and the other classes) indicate higher separability.

Overall, class separability follows the relation:  
\[
\texttt{no\_mass} \leftrightarrow \texttt{mass\_pos1} 
> \texttt{mass\_pos2} 
> (\texttt{mass\_pos3}, \texttt{mass\_pos4}),
\]
where “$>$” denotes increasing ease of discrimination (i.e., decreasing overlap and misclassification risk). \\

\begin{center}
\includegraphics[width=\linewidth,height=0.35\textheight,keepaspectratio]{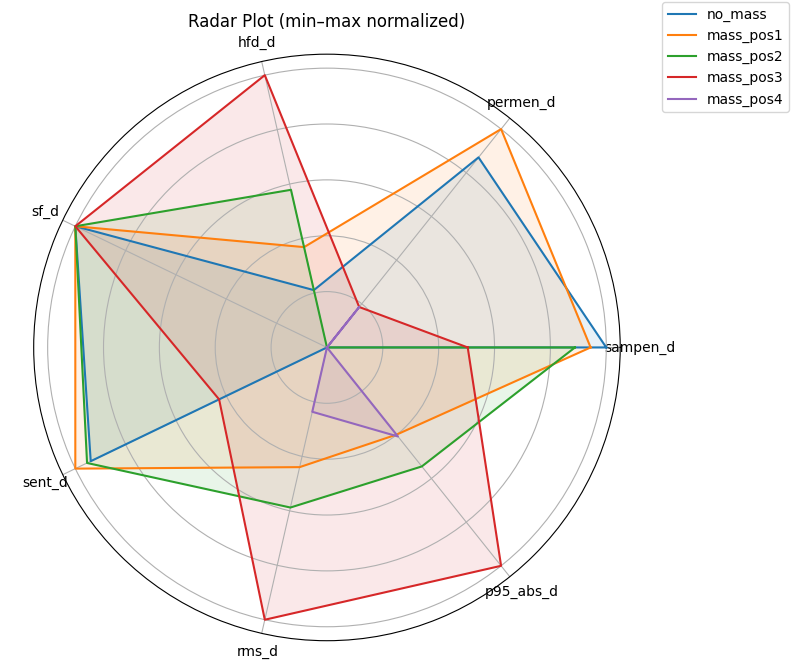}
\vspace{0.25em}
\small
\captionof{figure}{Larger pairwise overlaps (e.g., \texttt{no\_mass} vs. \texttt{mass\_pos1}) imply harder classification, while more distinct shapes (e.g., \texttt{mass\_pos3}, \texttt{mass\_pos4}) imply easier separability.}
\label{fig:delta_tau_classwise}
\end{center}

\subsection{Motivation}Directly feeding highly similar raw displacement signals into recurrent neural networks (RNNs) without reformulation or feature restructuring makes it difficult to achieve reliable classification performance. In our baseline study (Table 3), seven traditional models and seven recent deep learning models were applied directly to the original signals without any preprocessing. The results show that even recent RNN-based architectures failed to achieve sufficiently high accuracy under this setting.

This observation is consistent with recent studies reporting that raw displacement sequences often lead to poor classification performance due to long redundant sequences, misaligned temporal dynamics, and the difficulty of capturing long-term dependencies in sequential data \citep{he2024indepth, fang2024dam, ye2024light}. Although aggressive sequence shortening can reduce redundancy, it may also remove critical temporal transitions and frequency components, thereby degrading discriminative information and increasing misclassification rates \citep{ye2024light, hermiz2021impact, bao2022effect}.

To address these limitations, we propose a reformulation and preprocessing framework that segments the displacement signals into multiple physically meaningful components while preserving their overall dynamic characteristics. Specifically, the original sequences are reformatted into cumulative displacement states aligned at physically interpretable arrival times. Three descriptors are used for alignment: Best $\tau$, which captures the dominant time constant of the system; Knee $\tau$, which identifies the transition between fast and slow responses; and $S^{\star}$, which serves as a reliability measure to preserve trustworthy temporal patterns. This reformulation shortens sequence length, improves feature compactness, and enhances temporal alignment across signals, consistent with recent advances in time-series representation learning and RNN training.

\subsection{Optimal Sampling Interval Selection}
Determining an appropriate sampling interval $\Delta\tau$ for time-series alignment 
requires balancing spectral coverage with class discriminability. 
Overly dense sampling introduces redundancy and amplifies noise, leading to overfitting 
and unstable training, whereas excessively sparse sampling discards essential temporal 
transitions. 
According to \textit{compressed sensing theory}, when signals exhibit sparsity, 
sub-Nyquist sampling can preserve their essential structure while reducing data 
requirements~\cite{dai2018subnyquist}. 

For spectral guidance, the critical frequency $f^{\star}$ can be estimated from the 
power spectral density (PSD), with the corresponding Nyquist-based reference interval 
defined as $\Delta\tau_{\text{Nyq}} \approx \tfrac{1}{2 f^{\star}}$. 
Furthermore, theoretical studies suggest that when spectral support is unknown, 
stable reconstruction may require sampling at up to twice the Nyquist rate
~\cite{landau2009spectral}.

\subsubsection{Spectral Guideline}
For each class $c$, the power spectral density (PSD) $P_c(f)$ is estimated, and the 
critical frequency $f_c^\star$ is defined as the frequency below which 95\% of the 
signal energy is contained~\cite{oppenheim2010discrete}:
\begin{equation}
\int_{0}^{f_c^\star} P_c(f)\,df \;=\; 0.95 \int_{0}^{\infty} P_c(f)\,df .
\end{equation}
The overall reference frequency is then defined as
\begin{equation}
f^\star = \max_c f_c^\star, 
\quad 
\Delta\tau_{\text{Nyq}} \;\approx\; \frac{1}{2 f^\star},
\end{equation}

The admissible search space for the sampling interval $\Delta\tau$ is further 
constrained as
\begin{equation}
\Delta\tau \in [\beta \Delta\tau_{\text{Nyq}}, \gamma \Delta\tau_{\text{Nyq}}],
\end{equation}
with $\beta=0.5$ and $\gamma=3$ as typical values following 
standard signal processing guidelines~\cite{papoulis1977signal}.

\subsubsection{Data-Driven Selection: Discriminability vs. Redundancy}
For each candidate $\Delta\tau$, construct a uniform grid
\begin{equation}
G(\Delta\tau) = \{ \tau_j = j \Delta\tau \;|\; 0 \leq \tau_j \leq T \}, 
\quad M = |G|.
\end{equation}

\noindent
\textbf{(a) Discriminability:}  
The ANOVA $F$-score~\cite{fisher1925statistical} at each $\tau_j$ is  
\begin{equation}
F(\tau_j) = 
\frac{\sum_{c=1}^K N_c \left(\mu_c(\tau_j) - \mu(\tau_j)\right)^2}
{\sum_{c=1}^K \sum_{n \in D_c} \left(x_n^{(c)}(\tau_j) - \mu_c(\tau_j)\right)^2 + \epsilon}.
\end{equation}

\noindent
\textbf{(b) Redundancy Penalty:}  
The correlation-based redundancy penalty, as introduced by \cite{fisher1925statistical} and inspired by earlier work in sensor placement \cite{krause2008near}, is defined as:
\begin{equation}
R = \frac{2}{M(M-1)} \sum_{1 \leq i < j \leq M} r(\tau_i, \tau_j) 
\exp \left( - \frac{|\tau_i - \tau_j|}{\ell} \right).
\end{equation}
This penalty discourages selecting highly correlated parameters that are close in the domain, with \(\ell\) controlling the decay of influence.

\noindent
\textbf{(c) Combined Objective:}  
The final objective score, which balances performance, redundancy, and resource constraints, is given by:
\begin{equation}
S(\Delta\tau) = \frac{1}{M} \sum_{j=1}^M F(\tau_j) 
- \lambda_r R 
- \lambda_m \frac{M}{T}.
\end{equation}
Here, \(F(\tau_j)\) represents the utility of parameter \(\tau_j\), while \(\lambda_r\) and \(\lambda_m\) tune the trade-off between redundancy and cardinality.

\subsubsection{Final Selection}
The optimal interval is selected as:
\begin{equation}
\Delta\tau^\star = 
\arg\max_{\Delta\tau \in [\beta \Delta\tau_{\text{Nyq}}, \gamma \Delta\tau_{\text{Nyq}}]} 
\big\{ S(\Delta\tau) \;   \big\}.
\end{equation}

Unlike traditional sampling strategies that prioritize high-fidelity signal reconstruction, this optimization framework prioritizes \textit{information density}. The resulting optimal interval $\Delta\tau^\star$ represents a task-specific resolution that maximizes the neural network's capacity to learn class-distinguishing patterns while simultaneously minimizing the computational overhead associated with the input sequence length.


\subsection{Short-Time Fourier Transform (STFT)}

To analyze the time-varying spectral characteristics of signals, we employ the Short-Time Fourier Transform (STFT) based on the $\Delta\tau^\star$ interval, which provides a joint time-frequency representation by applying a sliding window across the signal \cite{cohen1995time}. Unlike the conventional Fourier Transform, which captures only global frequency content, the STFT localizes spectral content in time \cite{rioul1991wavelets}. 

The STFT of a continuous-time signal \(x(t)\) is defined as:
\begin{equation}
X(t, \omega) = \int_{-\infty}^{\infty} x(\tau) \, w(\tau - t) \, e^{-j \omega \tau} \, d\tau,
\end{equation}
where \(w(\tau - t)\) is a window function centered at time \(t\), and \(\omega\) denotes the angular frequency \cite{oppenheim1999discrete}. The magnitude \(|X(t, \omega)|\) reveals the evolution of frequency components, while the phase \(\angle X(t, \omega)\) provides complementary information.

For discrete-time signals \(x[n]\), the STFT is expressed as:
\begin{equation}
X(m, \omega) = \sum_{n=-\infty}^{\infty} x[n] \, w[n - m] \, e^{-j \omega n},
\end{equation}
where \(m\) represents the time shift or frame index \cite{oppenheim1999discrete}. This representation enables the extraction of temporal and spectral features critical for signal classification \cite{wang2022time}. We compute six features---dominant amplitude, sideband symmetry, second-peak offset, harmonic ratio, continuous wavelet transform (CWT), and CEEMDAN---over time intervals determined by the optimal \(\Delta \tau^\star\), as described in \cite{zhang2022adaptive}.

\begin{enumerate}
    \item \textbf{Dominant Amplitude:} 
    \begin{equation}
    z_1 = A_1,
    \end{equation}
    where \(A_1\) is the amplitude of the dominant frequency mode. This captures signal energy, useful when frequencies are close but energies differ \cite{lin2020vibration}.
    
    \item \textbf{Sideband Symmetry:}
    \begin{equation}
    E_L = \sum_{k_1-\Delta \le k < k_1} X[k]^2, \quad E_R = \sum_{k_1 < k \le k_1+\Delta} X[k]^2,
    \end{equation}
    \begin{equation}
    z_2 = S = \frac{|E_L - E_R|}{E_L + E_R}.
    \end{equation}
    This measures modulation symmetry, where asymmetry indicates faults or load changes \cite{}.
    
    \item \textbf{Second-Peak Offset:}
    \begin{equation}
    k_2 = \arg\max_{k \neq k_1} X[k], \quad z_3 = |f_{k_2} - f_{k_1}|,
    \end{equation}
    capturing harmonic or sideband spacing \cite{lin2020vibration}.
    
    \item \textbf{Harmonic Ratio:}
    \begin{equation}
    k_{2f} = \arg\min_k |f_k - 2f_1|, \quad z_4 = X[k_{2f}]/A_1,
    \end{equation}
    quantifying the strength of the second harmonic \cite{wang2022stft}.
    
    \item \textbf{Continuous Wavelet Transform (CWT):}
    \begin{equation}
    W_x(a,b) = \frac{1}{\sqrt{|a|}} \int_{-\infty}^{\infty} x(t) \, \psi^*\!\left(\frac{t-b}{a}\right) \, dt,
    \end{equation}
    \begin{equation}
    z_5 = \max_{a,b} \, |W_x(a,b)|,
    \end{equation}
    where \(x(t)\) is the input signal, \(\psi(t)\) is the mother wavelet, \(a\) is the scale, \(b\) is the translation, and \(^*\) denotes the complex conjugate \cite{rioul1991wavelets}. This captures localized time-frequency features.
    
    \item \textbf{CEEMDAN (Complete Ensemble Empirical Mode Decomposition with Adaptive Noise):}  
Signal \(x(t)\) is decomposed into intrinsic mode functions (IMFs) using noisy EMD realizations with adaptive noise.  
We extract the \emph{energy ratio feature}:
\begin{equation}
z_6 = \frac{\sum_{i=1}^{N} \|\text{IMF}_i(t)\|^2}{\|x(t)\|^2},
\end{equation}
where \(N\) is the number of IMFs. CEEMDAN improves stability and mitigates mode mixing compared to classical EMD~\cite{torres2011ceemdan}.

\end{enumerate}

Identifying signals that exhibit highly similar characteristics poses a significant challenge for deep learning models. In this chapter, we examine the properties of the signals collected from our experimental setup using a seven-dimensional feature vector. We then describe in detail our proposed methodology, which enables effective splitting of the signals while preserving their intrinsic characteristics, thus improving the performance of deep learning-based signal identification.

\section{Methods} 

This study proposes a hybrid spectro--temporal preprocessing framework that transforms raw displacement signals into structured representations for robust deep learning classification. 
As illustrated in Fig.2 and Table 1, the proposed pipeline consists of three main stages: 
\textit{Base Line Approach}, \textit{Spectro--Temporal Alignment (STA)}, and \textit{Hybrid Spectro--Temporal Fusion (HSTF)}.

\paragraph{Base Line Approach:}  
The base processing stage establishes baseline performance by evaluating ten representative machine learning models on the raw dataset without any preprocessing, providing a reference for assessing the effectiveness of the proposed framework.

\paragraph{Spectro--Temporal Alignment (STA):}
Given raw signals $(t_i, x_i)$, the power spectral density (PSD) of each class is first analyzed to estimate the effective bandwidth. 
Based on the Nyquist criterion, a reference sampling interval $\Delta\tau$ is determined and the signals are resampled accordingly. 
To avoid boundary distortion and unstable transient regions, a trimming margin $\alpha$ is applied. 
This stage preserves the physical characteristics of the original signals while reducing redundant samples.
The resampled signals are segmented into uniform temporal windows with interval $\Delta\tau$, producing aligned cumulative displacement states $x^{(m)}$. 
This windowing process mitigates temporal misalignment across different experimental conditions and generates structured sequences suitable for sequential deep learning models. 
For each window, a short-time Fourier transform (STFT) is applied to obtain the spectral map $S^{(m)}(\omega_k)$.

\paragraph{Hybrid Spectro--Temporal Fusion (HSTF):}
From each spectral map, discriminative spectral features $z^{(m)}$ are extracted, including dominant amplitude, sideband symmetry, second-peak offset, and harmonic ratio (optionally CWT and CEEMDAN-based features). 
The temporal representation $x^{(m)}$ and the spectral feature vector $z^{(m)}$ are then concatenated to form stacked per-window tokens $[x^{(m)} \| z^{(m)}]$. 
Finally, all tokens are stacked to construct the fused sequence representation $\mathbf{Z}_i$, which is used as the input to the deep learning classifier.


\begin{figure*}[t]  
\centering

\begin{minipage}{\textwidth}
\centering
\footnotesize
\setlength{\tabcolsep}{4pt}
\renewcommand{\arraystretch}{1.1}
\begin{tabularx}{\textwidth}{@{} p{0.15\textwidth} p{0.18\textwidth} p{0.22\textwidth} X @{}}
\toprule
\textbf{Level} & \textbf{Symbol} & \textbf{Shape / Indexing} & \textbf{Description} \\
\midrule
Raw signals & $t_i$ & $\mathbb{R}_+^{N_i}$ & Time samples ($i=0\ldots4$). \\
& $x_i$ & $\mathbb{R}^{N_i}$ & Displacement sequence aligned to $t_i$. \\
& $y_i$ & $\{0,1,2,3,4\}$ & Class label (five conditions). \\[0.3mm]
Resampling & $\Delta t$ & scalar & Uniform Nyquist-safe step. \\
& $\alpha$ & scalar & Safety margin (edge trimming). \\[0.3mm]
Windowing & $x^{(m)}$ & $\mathbb{R}^{L}$ & $m=1,\ldots,M_i$ windows of length $L$. \\
& $t^{(m)}$ & $\mathbb{R}^{L}$ & Time indices per window. \\[0.3mm]
Spectrum & $w_k$ & $k=0,\ldots,99$ & Frequency bins (STFT/CWT). \\
& $S^{(m)}(w_k)$ & $\mathbb{R}$ & Spectral magnitude per bin. \\[0.3mm]
Features & $\mathbf z^{(m)}$ & $\mathbb{R}^{6}$ &
$[z_1\!\ldots\!z_6]^\top$: dominant amplitude, sideband symmetry,
second-peak offset, harmonic ratio, CWT energy, CEEMDAN index. \\[0.3mm]
Window stack & $\mathbf X_i$ & $\mathbb{R}^{M_i\times L}$ & Rows are $x^{(m)}$ for class $i$. \\
Feature matrix & $\mathbf Z_i$ & $\mathbb{R}^{M_i\times 6}$ & Rows are $\mathbf z^{(m)}$ for class $i$. \\
\textbf{Stacked representation} & $\hat{\mathbf Z}_i=[\mathbf X_i \,|\, \mathbf Z_i]$
& $\mathbb{R}^{M_i\times (L+6)}$ & Column-wise concatenation per window. \\
Learning set & $\hat{\mathcal Z}$ & $\bigcup_i \{(\hat{\mathbf Z}_i, y_i)\}$ & All stacked matrices with labels. \\
\bottomrule
\end{tabularx}
\captionof{table}{Data organization and re-formatting from raw displacement signals
to stacked per-window representations that concatenate samples $x^{(m)}$
with features $\mathbf z^{(m)}$.}
\label{tab:data-format}
\end{minipage}

\vspace{2mm}

\begin{minipage}{\textwidth}
\centering
\resizebox{\textwidth}{!}{%
\begin{tikzpicture}[
  node distance=12mm,
  box/.style={draw, rounded corners=2mm, inner sep=3mm, align=center, font=\footnotesize},
  badge/.style={draw, rounded corners=3mm, inner sep=2.5mm, minimum width=22mm, align=center, font=\bfseries},
  >={Latex[length=2mm]}, thick]
\node[box] (raw) {Raw signals\\$(t_i, x_i)$};
\node[box, right=12mm of raw] (resamp) {Resample to $\Delta t$\\Trim margin $\alpha$};
\node[box, right=12mm of resamp] (win) {\textbf{Windowing}\\$x^{(m)}\!\in\!\mathbb{R}^{L}$};
\node[box, right=12mm of win] (spec) {Spectral map\\$S^{(m)}(w_k)$};
\node[box, right=12mm of spec] (feat) {Feature extractor $\Phi$:\\$x^{(m)}\!\mapsto\!\mathbf z^{(m)}$};
\node[box, right=12mm of feat] (stack) {\textbf{Stacked per-window tokens}\\$[x^{(m)} \,|\, \mathbf z^{(m)}] \Rightarrow \hat{\mathbf Z}_i$};
\draw[->] (raw) -- (resamp);
\draw[->] (resamp) -- (win);
\draw[->] (win) -- (spec);
\draw[->] (spec) -- (feat);
\draw[->] (feat) -- (stack);
\node[badge, above=8mm of raw]   (basebadge) {Base};
\node[badge, above=8mm of win]   (stabadge) {STA};
\node[badge, above=8mm of stack] (hstfbadge) {HSTF};
\draw[->, line width=0.5pt] (raw.north)   -- ++(0,1mm) -- (basebadge.south);
\draw[->, line width=0.5pt] (win.north)   -- ++(0,1mm) -- (stabadge.south);
\draw[->, line width=0.5pt] (stack.north) -- ++(0,1mm) -- (hstfbadge.south);
\end{tikzpicture}}
\captionof{figure}{Pipeline from raw signals to stacked per-window representation
$[x^{(m)} \,|\, \mathbf z^{(m)}]\Rightarrow \hat{\mathbf Z}_i$.}
\label{fig:reformat-pipeline}
\end{minipage}
\end{figure*}

\subsection{Deep Learning Model for Categorical classification }
Let $\mathcal{D}$ denote a domain consisting of vibration-induced displacement signals collected from a single device under a fixed experimental configuration. The signals are obtained from a shaker-excited PLA cantilever beam with an attached lumped mass to emulate defects, and the tip displacement is measured using a laser displacement sensor. Thus, each domain represents responses acquired under identical physical and sensing conditions.

\[
\mathcal{D} = \{ \eta, \delta_{1}, \delta_{2}, \delta_{3}, \delta_{4} \},
\]
where $\eta$ represents the healthy state and $\delta_{k}$, $k=1,2,3,4$, correspond to different levels of structural damage. Each signal is a time--displacement sequence
\[
x = \{ (t_j, u_j) \}_{j=1}^{T},
\]
where $t_j \in \mathbb{R}$ denotes the time index, $u_j \in \mathbb{R}$ the corresponding displacement, and $T$ the sequence length. Each condition state is then described as a collection of such signals:
\[
\eta = \Big\{ x_i^{(\eta)} \Big\}_{i=1}^{n_\eta}, \qquad 
\delta_k = \Big\{ x_i^{(\delta_k)} \Big\}_{i=1}^{n_{\delta_k}}, \quad k=1,2,3,4.
\] 

The learning task is to construct a classifier
\[
f : \mathcal{X} \rightarrow \mathcal{D},
\]
which maps an input signal $x \in \mathcal{X}$ to its corresponding condition label $y \in \mathcal{D}$. 
Formally, given training data 
\[
\mathcal{T} = \{(x_i, y_i)\}_{i=1}^{N}, \quad y_i \in \mathcal{D},
\]
the objective is to minimize the empirical risk
\[
\min_{f \in \mathcal{F}} \; \frac{1}{N} \sum_{i=1}^N \ell \big(f(x_i), y_i \big),
\]
where $\ell(\cdot,\cdot)$ is a suitable loss function (e.g., cross-entropy) and $\mathcal{F}$ is the hypothesis space of candidate classifiers.

\subsection{Spectro-Temporal Alignment (STA)}

To ensure uniform temporal representation across vibration-harvested signals, the \textbf{Spectro-Temporal Alignment (STA)} module establishes a data-driven sampling interval and segmentation procedure prior to deep learning.

Let each raw displacement sequence be denoted as
\[
x_c(t) = \{ (t_j, u_j) \}_{j=1}^{T_c}, \quad c \in \{1, \dots, C\},
\]
where $t_j$ represents the time index and $u_j$ the measured displacement.
The STA process begins by \textit{resampling} each $x_c(t)$ to a uniform sampling interval $\Delta\tau$, trimming boundary margins by $\alpha$ to suppress transient edge effects.

To determine the most informative interval, a bounded score $S(\Delta\tau_i)\in[0,1]$ is evaluated across all Nyquist-aligned candidate lags $\{\Delta\tau_i\}_{i=1}^{N_\tau}$. 
This score jointly captures three complementary aspects: 
(1) \textit{class separability}, quantified using the ANOVA $F$-statistic~\citep{ghosh2021anova}; 
(2) \textit{temporal redundancy}, penalized through correlation-based dependency measures~\citep{li2023adaptive}; and 
(3) \textit{sampling complexity}, which balances model efficiency with spectral fidelity~\citep{sharma2022spectral}. 
Together, these criteria ensure that the chosen interval maximizes discriminative information while minimizing redundant or oversampled representations.

\begin{equation}
\Delta\tau^\star = \arg\max_i S(\Delta\tau_i),
\end{equation}
while the \textit{knee point} $\tau_k$ identifies the onset of diminishing returns in temporal resolution.

Once $\Delta\tau^\star$ is fixed, each resampled sequence is \textit{windowed} into overlapping segments
\[
x^{(m)} \in \mathbb{R}^L, \quad m = 1, \dots, M,
\]
forming a batch-ready tensor representation for deep learning.
This alignment ensures that all input windows are temporally synchronized, spectrally balanced, and comparable across classes—allowing the subsequent network to focus on discriminative structural dynamics rather than sampling irregularities.

\subsection{Hybrid spectro–temporal Fusion(HSTF)}
\label{sec:Proposed Approach}
The HSTF operates in conjunction with the neural network–based classification described in Subsection 3.4. It segments the deep learning input sequence into designated Short-Time Analyses (STA) and, at each segment, combines the six feature extraction values introduced in Section 2.2 with the corresponding portion of the original signal.

Each displacement signal $x = \{(t_j, u_j)\}_{j=1}^T$ is first segmented 
using a sliding window of size $w$, producing subsequences
\[
x^{(m)} = \{(t_j, u_j)\}_{j = (m-1)w+1}^{mw}, 
\quad m = 1, 2, \dots, M,
\]
where $M = \lfloor T / w \rfloor$ is the total number of windows.  \\

Each subsequence $x^{(m)}$ is then transformed into a 
feature vector by a feature extraction operator 
$\Phi: x^{(m)} \mapsto \mathbf{z}^{(m)} \in \mathbb{R}^{10}$, 
where 
\begin{equation}
    \mathbf{z}^{(m)} = 
    \big[z_{1}^{(m)}, z_{2}^{(m)}, \ldots, z_{10}^{(m)}\big]^{\top},
    \quad m = 1,2,\ldots,M .
\end{equation} Here, $z_{i}^{(m)}$ denotes the $i$-th spectral features z1, z2, ..., z6 computed from the $m$-th window \textbf{(Appendix Fig. 6-8)}. 
Thus, the complete signal $x$ is represented by a collection of 
feature vectors
\begin{equation}
    \mathcal{Z} = \{\mathbf{z}^{(m)}\}_{m=1}^{M}, 
    \quad \mathbf{z}^{(m)} \in \mathbb{R}^{6}.
\end{equation}

\begin{center}
  \includegraphics[width=\linewidth,height=0.35\textheight,keepaspectratio]{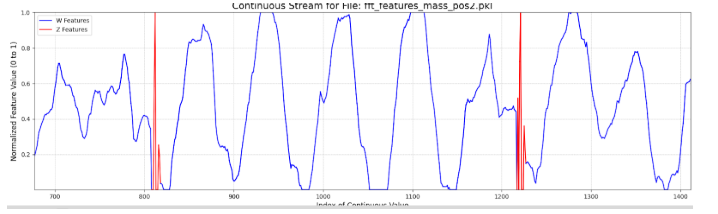}
  \vspace{0.25em}
  \small
  \captionof{figure}{Visualization of the healthy-state signal representation. The blue curve shows the original displacement signal 
  $\eta = \{ x_i^{(\eta)} \}_{i=1}^{n_\eta}$, while the red segments 
  highlight its transformed representation 
  $\eta = \{ \widetilde{x}_i^{(\eta)} \}_{i=1}^{n_\eta}$, obtained by 
  segmenting the signal into windows and mapping each subsequence to a feature vector.}
  \label{fig:signals}
\end{center}

For each displacement signal $x = \{(t_j,u_j)\}_{j=1}^T$, the segmentation step produces subsequences $\{x^{(m)}\}_{m=1}^M$. Each subsequence is then associated with its feature vector $\mathbf{z}^{(m)} \in \mathbb{R}^{6}$, and the collection of features is denoted by 
$\mathcal{Z} = \{\mathbf{z}^{(m)}\}_{m=1}^M$. 

Thus, the representation of the original signal $x$ is augmented as 
\begin{equation}
    \widetilde{x} \;=\; \Big\{ \big(x^{(m)},\, \mathbf{z}^{(m)}\big) \Big\}_{m=1}^M,
\end{equation}
where each pair $\big(x^{(m)}, \mathbf{z}^{(m)}\big)$ consists of the 
raw subsequence and its corresponding feature vector. 

For a dataset containing $N$ signals 
$\{x_i\}_{i=1}^N$, the complete augmented representation is given by 
\begin{equation}
    \widetilde{\mathcal{X}} 
    = \Big\{ \widetilde{x}_i \Big\}_{i=1}^N
    = \Big\{ \big(x_i^{(m)}, \mathbf{z}_i^{(m)}\big) \Big\}_{i=1,m=1}^{N,\;M_i},
\end{equation}
where $M_i = \lfloor T_i / w \rfloor$ is the number of windows 
for the $i$-th signal. 
After segmentation and feature extraction, each signal 
$x_i^{(\eta)}$ or $x_i^{(\delta_k)}$ is transformed into an 
augmented representation
\[
    \widetilde{x}_i^{(\cdot)} 
    = \Big\{ \big(x_i^{(m)}, \mathbf{z}_i^{(m)}\big) \Big\}_{m=1}^{M_i},
\]
where $x_i^{(m)}$ denotes the $m$-th subsequence of the $i$-th signal, 
and $\mathbf{z}_i^{(m)} \in \mathbb{R}^{6}$ is the corresponding 
feature vector. 

Thus, the original datasets
\[
    \eta = \Big\{ x_i^{(\eta)} \Big\}_{i=1}^{n_\eta}, 
    \qquad
    \delta_k = \Big\{ x_i^{(\delta_k)} \Big\}_{i=1}^{n_{\delta_k}}, 
    \quad k=1,2,3,4
\]
are transformed into the augmented datasets
\[
    \widetilde{\eta} = 
    \Big\{ \widetilde{x}_i^{(\eta)} \Big\}_{i=1}^{n_\eta}, 
    \qquad
    \widetilde{\delta}_k = 
    \Big\{ \widetilde{x}_i^{(\delta_k)} \Big\}_{i=1}^{n_{\delta_k}}, 
    \quad k=1,2,3,4,
\]
where each $\widetilde{x}_i^{(\cdot)}$ contains both the 
time–domain subsequences and their associated feature vectors. The following figure shows the transformed signal as an example.

\subsection{Neural Network-Based Classification}

We consider the five-class signal classification problem, where each signal 
sequence belongs to one of the experimental cases 
($\eta, \delta_{1}, \delta_{2}, \delta_{3}, \delta_{4}$).  
The training dataset is given by
\begin{equation}
\{(Z_i, y_i)\}_{i=1}^{N}, \quad y_i \in \{0,1,2,3,4\},
\end{equation}
where $Z_i = \{ z_i^{(m)} \}_{m=1}^{M_i}$ represents the extracted feature sequence for the $i$-th sample, and $y_i$ is its ground-truth class label. 

We have applied a comprehensive set of 14 algorithms spanning both deep learning 
and traditional machine learning approaches. 
The deep learning models include Bi-GRU~\citep{schuster1997bigru}, 
Bi-LSTM~\citep{hochreiter1997lstm}, 
CNN~\citep{lecun1998cnn}, 
GRU~\citep{cho2014gru}, 
LSTM~\citep{hochreiter1997lstm}, 
TCN~\citep{bai2018tcn}, 
and Transformer~\citep{vaswani2017transformer}. 
For comparison, we also implemented classical machine learning methods, namely 
Decision Tree~\citep{quinlan1993c45}, 
Gaussian Na{\"i}ve Bayes (NB)~\citep{russell2010aibook}, 
K-Nearest Neighbors (KNN)~\citep{russell2010aibook}, 
LightGBM~\citep{ke2017lightgbm}, 
Logistic Regression (Log. Reg.)~\citep{cox1958logreg}, 
Random Forest~\citep{breiman2001rf}, 
and Support Vector Machine (SVM)~\citep{cortes1995svm}. 
Among these, we provide a detailed explanation of the 
LSTM (Long Short-Term Memory)~\citep{hochreiter1997lstm} case to illustrate the methodology.

\subsubsection{Sequence Encoding via LSTM}
Each feature sequence $Z_i$ is passed through an LSTM encoder parameterized 
by $\theta_{\mathrm{enc}}$:
\begin{equation}
\{ h_m \}_{m=1}^{M_i} = \mathrm{LSTM}(Z_i; \theta_{\mathrm{enc}}),
\end{equation}
where $h_m$ denotes the hidden state at time step $m$.  
To obtain a fixed-dimensional representation, we use either the last hidden state
\begin{equation}
s_i = h_{M_i},
\end{equation}
or an attention-pooled representation
\begin{equation}
s_i = \sum_{m=1}^{M_i} \alpha_m h_m, \quad 
\alpha_m = \frac{\exp(u_m)}{\sum_{j=1}^{M_i} \exp(u_j)},
\end{equation}
where $\alpha_m$ are normalized attention weights.

\subsubsection{Projection to Class Space}
The sequence embedding $s_i \in \mathbb{R}^d$ is projected into a 
five-dimensional logit vector:
\begin{equation}
\ell_i = W_o s_i + b_o, \quad \ell_i \in \mathbb{R}^{5},
\end{equation}
where $W_o \in \mathbb{R}^{5 \times d}$ and $b_o \in \mathbb{R}^5$.

\subsubsection{Softmax Probability Distribution}
Class probabilities are obtained via the softmax function:
\begin{equation}
p_i^{(c)} = \frac{\exp(\ell_i^{(c)})}{\sum_{k=0}^{4} \exp(\ell_i^{(k)})},
\quad c = 0,\ldots,4.
\end{equation}

\subsubsection{Training Objective}
The classification loss for sample $i$ is the cross-entropy:
\begin{equation}
L_i = - \sum_{c=0}^{4} y_i^{(c)} \log p_i^{(c)},
\end{equation}
where $y_i^{(c)}$ is a one-hot encoding of the true class.  
The overall objective with $\ell_2$ regularization is
\begin{equation}
\mathcal{L}(\theta) = \frac{1}{N} \sum_{i=1}^{N} L_i 
+ \lambda \lVert \theta \rVert_2^2,
\end{equation}
where $\lambda$ controls the weight decay.

\subsubsection{Optimization and Inference}
The model parameters $\theta = \{\theta_{\mathrm{enc}}, W_o, b_o\}$ 
are optimized via stochastic gradient descent or Adam:
\begin{equation}
\theta \leftarrow \theta - \eta \, \nabla_\theta \mathcal{L}(\theta),
\end{equation}
with learning rate $\eta$.  
At inference time, the predicted class label is obtained as
\begin{equation}
\hat{y}_i = \arg\max_{c \in \{0,1,2,3,4\}} \; p_i^{(c)}.
\end{equation}

\begin{algorithm}[t]
\caption{Segmentation, Feature Extraction, and Dataset Augmentation}
\begin{algorithmic}[1]
\Require $\eta,\ \{\delta_k\}_{k=1}^4,\ w,\ \Phi$
\Ensure $\widetilde{\eta},\ \{\widetilde{\delta}_k\},\ \{\mathcal{Z}_i\}$
\ForAll{$C \in \{\eta,\delta_1,\delta_2,\delta_3,\delta_4\}$}
  \ForAll{signals $x_i^{(C)}=\{(t_j,u_j)\}_{j=1}^{T_i}$}
    \State $M_i \gets \left\lfloor T_i/w \right\rfloor$
    \State $\widetilde{x}_i^{(C)} \gets \emptyset$, \quad $\mathcal{Z}_i \gets \emptyset$
    \For{$m=1$ \textbf{to} $M_i$}
      \State $x_i^{(m)} \gets \{(t_j,u_j)\}_{j=(m-1)w+1}^{mw}$
      \State $\mathbf{z}_i^{(m)} \gets \Phi\!\big(x_i^{(m)}\big)$
      \State append $\big(x_i^{(m)},\mathbf{z}_i^{(m)}\big)$ to $\widetilde{x}_i^{(C)}$
      \State append $\mathbf{z}_i^{(m)}$ to $\mathcal{Z}_i$
    \EndFor
    \State append $\widetilde{x}_i^{(C)}$ to $\widetilde{C}$
  \EndFor
\EndFor
\State \textbf{return} $\widetilde{\eta},\,\widetilde{\delta}_1,\dots,\widetilde{\delta}_4$
\end{algorithmic}
\end{algorithm}

\begin{algorithm}[t]
\caption{Five-Class Signal Classification with LSTM}
\label{alg:lstm-classification}
\begin{algorithmic}[1]
\Require Training dataset $\{ (Z_i, y_i) \}_{i=1}^{N}$, with 
$Z_i = \{ z_i^{(m)} \}_{m=1}^{M_i}$, $y_i \in \{0,1,2,3,4\}$; 
LSTM parameters $\theta_{\mathrm{enc}}$; classifier parameters $(W_o, b_o)$.
\Ensure Trained parameters $\theta = \{ \theta_{\mathrm{enc}}, W_o, b_o \}$; predictions $\{ \hat{y}_i \}$.

\For{epoch $=1$ to $E$}
    \For{each sequence $Z_i$}
        \State \textbf{Encoding:} $\{ h_m \}_{m=1}^{M_i} \gets \mathrm{LSTM}(Z_i; \theta_{\mathrm{enc}})$
        \State Obtain representation $s_i \gets h_{M_i}$ \hfill (last hidden state or attention pooling)
        \State \textbf{Projection:} $\ell_i \gets W_o s_i + b_o \in \mathbb{R}^5$
        \State \textbf{Softmax:} $p_i^{(c)} \gets \frac{\exp(\ell_i^{(c)})}{\sum_{k=0}^4 \exp(\ell_i^{(k)})}$
        \State \textbf{Loss:} $L_i \gets - \sum_{c=0}^4 y_i^{(c)} \log p_i^{(c)}$
    \EndFor
    \State \textbf{Parameter Update:} 
    $\theta \gets \theta - \eta \, \nabla_\theta \Big( \frac{1}{N} \sum_{i=1}^N L_i + \lambda \lVert \theta \rVert_2^2 \Big)$
\EndFor

\For{$i = 1,\ldots,N$}
    \State \textbf{Prediction:} $\hat{y}_i \gets \arg\max_{c \in \{0,1,2,3,4\}} p_i^{(c)}$
\EndFor

\State \Return $\theta, \{ \hat{y}_i \}$
\end{algorithmic}
\end{algorithm}


\section{Experimental Results}

\subsection{Device Setup}
A laser displacement sensor (optoNCDT 1420, Micro-Epsilon) was employed to measure the time response of a 3D-printed PLA cantilever beam (108~mm~\texttimes~25~mm~\texttimes~3~mm) subjected to random vibration excitation. The beam's response was sampled at a rate of 4~kHz, while the random input from the shaker was measured via an accelerometer (352C33, PCB Piezotronics) mounted on the shaker, with a sampling rate of 1~kHz.

\subsection{Implementation Details}
All experiments were conducted on a Lambda Workstation connected to a local area network 
with a 1000 Mbps network speed. The workstation was equipped with two NVIDIA RTX 3090 GPUs, 
an Intel Core i9-10980XE CPU (18 cores), 256 GB of memory, and ran on Ubuntu 12.04. 
For application development, we employed the PyCharm Integrated Development Environment (IDE), 
which provides seamless integration with the \texttt{pip3} utility required for Python-based development.

\begin{center}
  \includegraphics[width=0.7\linewidth,height=0.35\textheight,keepaspectratio]{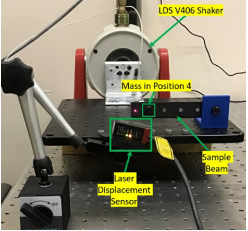}
  \vspace{0.25em}
  \small

  \captionsetup[figure]{
    justification=justified,
    singlelinecheck=false
  }

  \captionof{figure}{Experimental setup for vibration testing. The PLA cantilever beam (right) is mounted on a fixture driven by an LDS V406 shaker (background). A 7.4\,g mass is installed in \emph{Position~4} to emulate a defect and shift resonance. A laser displacement sensor (optoNCDT~1420) measures tip displacement, while the shaker input is monitored via an accelerometer mounted on the shaker (not shown in labels).}
  \label{fig:setup}
\end{center}

\subsection{Dataset Generation}
A test beam was fabricated using additive manufacturing with a vertical orientation, 0.2~mm layer height, and 100\% infill density to ensure structural uniformity. To introduce controlled defect conditions, four holes of 6.25~mm diameter were drilled along the beam at intervals of approximately 26~mm, and in separate experiments, a 7.4~g mass was placed in each hole to emulate localized stiffness and mass variations, generating four defect cases ($\delta_{1}$--$\delta_{4}$). Together with the pristine beam ($\eta$), five test cases were considered in total. The beam was excited by a shaker under broadband random vibration, while a laser displacement sensor (4~kHz) measured the beam tip displacement and an accelerometer (1~kHz) recorded the shaker input. Each trial lasted about 10~s, yielding multiple signal segments that were processed in both time and displacement domains to extract dynamic features sensitive to defect conditions, with representative plots provided in Fig. 13 in Appendix.

\subsection{Baseline Approach (Original Data Set)}
During the baseline evaluation, the original dataset was tested without any preprocessing or alignment. The details of the experimental setup used for running the code are summarized in Table I. 
Both traditional machine learning models and recent deep learning approaches were applied directly to the raw signals, providing a reference point against which the accuracy gains of the proposed Spectro-Temporal methods could be measured.

In all evaluations, five-fold cross-validation ($k=5$) was employed. 
The choice of five folds reflects a balance between computational efficiency and reliable estimation 
of generalization performance, ensuring that every subset of the dataset was systematically used 
for both training and validation.

\begin{table}[H]
\centering
\caption{Experimental Setup Parameters}
\begin{tabular}{|l|c|p{3cm}|}
\hline
\textbf{Parameter} & \textbf{Value} & \textbf{Explanation} \\ \hline
Timesteps          & 24    & Each input sequence spans 24 time points. \\ \hline
$n\_splits$        & 5      & Number of cross-validation folds . \\ \hline
win\_dur\_ratio    & 0.04   & Window duration ratio r. \\ \hline
sampling\_ratio    & 0.3    & 20\% of the available data. \\ \hline
start\_row\_index  & 4000   & where sequence extraction begins. \\ \hline
\end{tabular}
\end{table}

Interestingly, while recent deep learning models (e.g., Bi-GRU, Bi-LSTM, CNN, TCN) 
exhibited poor predictive performance on the original dataset, several traditional classifiers 
(e.g., RF, LGBM) achieved consistently high accuracy, Macro-F1, and Macro-AUC 
($\geq 0.95$), suggesting a dataset bias or representational inefficiency that hinders deep 
architectures but favors simpler models.
 
\begin{table}[htbp]
\centering
\caption{\textbf{Baseline} - Original Dataset (5-fold). Rows in bold indicate models achieving strong performance across all metrics (Accuracy, Macro-F1, Macro-AUC $\geq 0.95$).}
\scriptsize
\begin{tabular}{lccc}
\hline
\textbf{Model} & \textbf{Accuracy} & \textbf{Macro-F1} & \textbf{Macro-AUC} \\
\hline
Bi-GRU        & 0.442 $\pm$ 0.022 & 0.350 $\pm$ 0.030 & 0.766 $\pm$ 0.010 \\
Bi-LSTM       & 0.436 $\pm$ 0.030 & 0.317 $\pm$ 0.049 & 0.775 $\pm$ 0.042 \\
CNN           & 0.432 $\pm$ 0.011 & 0.361 $\pm$ 0.007 & 0.723 $\pm$ 0.011 \\
GRU           & 0.421 $\pm$ 0.022 & 0.299 $\pm$ 0.063 & 0.764 $\pm$ 0.028 \\
LSTM          & 0.416 $\pm$ 0.027 & 0.315 $\pm$ 0.025 & 0.739 $\pm$ 0.043 \\
TCN           & 0.427 $\pm$ 0.017 & 0.327 $\pm$ 0.027 & 0.766 $\pm$ 0.005 \\
Trans.        & 0.508 $\pm$ 0.032 & 0.423 $\pm$ 0.038 & 0.832 $\pm$ 0.035 \\
\midrule
DT            & 0.859 $\pm$ 0.012 & 0.858 $\pm$ 0.012 & 0.912 $\pm$ 0.007 \\
GNB           & 0.776 $\pm$ 0.017 & 0.775 $\pm$ 0.018 & 0.946 $\pm$ 0.005 \\
KNN           & 0.906 $\pm$ 0.004 & 0.906 $\pm$ 0.004 & 0.988 $\pm$ 0.001 \\
\textbf{LGBM} & \textbf{0.958 $\pm$ 0.004} & \textbf{0.958 $\pm$ 0.004} & \textbf{0.998 $\pm$ 0.001} \\
LR            & 0.820 $\pm$ 0.013 & 0.819 $\pm$ 0.013 & 0.961 $\pm$ 0.003 \\
\textbf{RF}   & \textbf{0.954 $\pm$ 0.005} & \textbf{0.954 $\pm$ 0.005} & \textbf{0.998 $\pm$ 0.000} \\
SVM           & 0.795 $\pm$ 0.009 & 0.784 $\pm$ 0.009 & 0.962 $\pm$ 0.003 \\
\hline
\end{tabular}
\end{table}

\subsection{Optimal Interval Selection ($\Delta \tau$)}
Accurate selection of the sampling interval ($\Delta \tau$) is crucial for time–series signal classification because it governs the balance between capturing informative temporal variations and avoiding redundancy from oversampling. If $\Delta \tau$ is too small, the signal features become highly redundant, inflating computational cost without improving discriminative power. Conversely, if $\Delta \tau$ is too large, important transient features may be lost, leading to poor classification performance. To address this, we systematically analyzed the objective function $S(\Delta \tau)$ across different classes, identifying class-specific trade-offs between temporal resolution and feature relevance. By comparing “best” (accuracy-maximizing) and “knee” (efficiency-driven) interval choices, we derived common optimal intervals that robustly preserve the informative classes ($\delta_1$–$\delta_3$) while minimizing redundancy, as illustrated in Fig. 5, 6 and summarized in Table 4. \\

In Fig. 5, Objective function $S(\Delta\tau)$ across classes healty and damaged ($\mu, \delta_1, \delta_2, \delta_3, \delta_4$). Each curve depicts class-specific performance as a function of $\Delta\tau$, 
with vertical dashed lines marking the estimated optimal values. 
The plot highlights how different classes respond to temporal resolution, 
revealing varying trade-offs between discriminative gain and redundancy.  \textbf{Fig.~6.} Classification accuracy ($S^\ast$) versus optimal lag ($\tau$). 

Best lag ($\tau^\ast$) values appear scattered across larger $\tau$, 
while knee lag ($\tau_k$) values cluster tightly at smaller $\tau$, 
indicating that $\tau_k$ yields a more compact and stable characterization 
of time-dependent dynamics.

\noindent
Table~III summarizes the optimal time interval parameters ($\tau$) and the resulting 
classification accuracy ($S^\ast$) for the multi-class signal dataset. The most 
informative classes, $\delta_2$--$\delta_4$, yield the highest accuracies 
($S^\ast \approx 0.52$--0.58) with $\tau^{\text{best}}$ values in the range of 
18--20\,ms. In contrast, $\mu$ and $\delta_1$ achieve comparatively lower accuracies 
($S^\ast \approx 0.49$--0.51), despite similar time lag parameters. These findings 
indicate that compact intervals around 18--20\,ms provide stable discriminative 
performance, while class $\mu$ contributes less effectively to the overall 
classification.

\begin{center}
  \includegraphics[width=\linewidth,height=0.35\textheight,keepaspectratio]{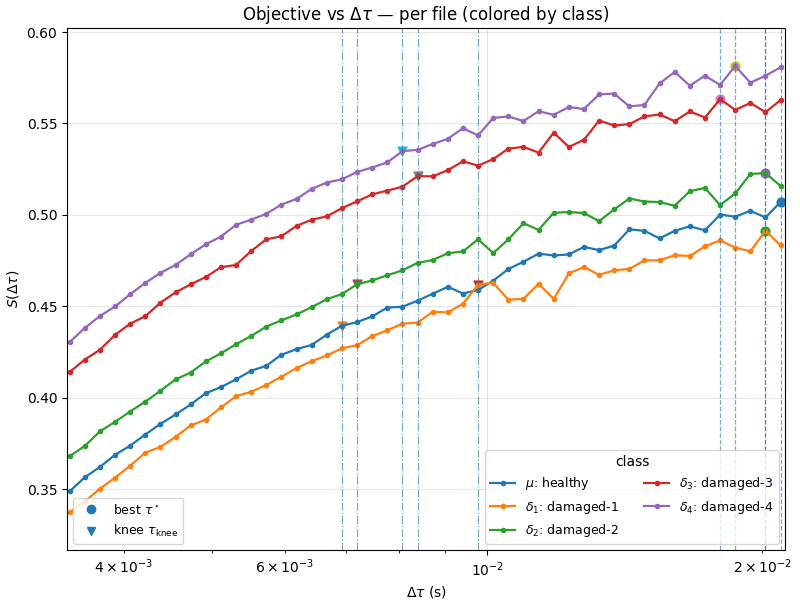}
  \vspace{0.25em}
 
    \small
    \captionsetup[figure]{
    justification=justified,
    singlelinecheck=false
  }
  \captionof{figure}{{Objective function $S(\Delta\tau)$ across healthy and damaged classes ($\mu$, $\delta_1$–$\delta_4$), showing class-specific performance and estimated optimal time intervals.} }
  \label{fig:signals}
\end{center}


\begin{table}[t]
\centering
\caption{Time Interval Parameters (\(\tau\)) and Accuracy (\(S^\star\)) for Multi-Class Signal Classification.}
\scriptsize
\begin{tabular}{lcccccc}
\toprule
\textbf{Label} & \textbf{Nyquist\_dt\_s} & \textbf{Best\_tau\_s} &  \textbf{Knee\_tau\_s} & \textbf{\(S^\star\)} \\
\midrule
\(\mu\)        & 0.007 & 0.021  & \textbf{0.007} & 0.507 \\
\(\delta_1\)   & 0.007 & 0.020  & \textbf{0.010} & 0.491 \\
\(\delta_2\)   & 0.007 & 0.020 & \textbf{0.007} & 0.523 \\
\(\delta_3\)   & 0.007 & 0.018  & \textbf{0.008} & 0.563 \\
\(\delta_4\)   & 0.007 & 0.019  & \textbf{0.008} & 0.581 \\
\bottomrule
\end{tabular}
\label{tab:delta_tau_summary}
\end{table}

\begingroup
\setlength{\abovedisplayskip}{4pt}
\setlength{\belowdisplayskip}{4pt}
\setlength{\abovedisplayshortskip}{2pt}
\setlength{\belowdisplayshortskip}{2pt}

Let $\mathcal{C}={\mu,\delta_1,\delta_2,\delta_3,\delta_4}$ denote all classes.
For $c\in\mathcal{C}$, let $\tau^{\text{best}}_c$ be the accuracy–maximizing interval ($\mathrm{s}$),
$S_c^\ast$ the corresponding accuracy, and $\tau^{\text{knee}}_c$ the knee time.

\noindent\textbf{(1) Accuracy-driven (Best-$\tau$).}
Define the $S^\ast$–weighted common interval
\begin{equation}
\tau^{\text{common}}*{\text{best}}
=\frac{\sum*{c\in\mathcal{C}} S_c^\ast,\tau^{\text{best}}*c}
{\sum*{c\in\mathcal{C}} S_c^\ast}.
\end{equation}

With
\[
\begin{aligned}
(\tau^{\text{best}}_\mu,S^\ast_\mu)&=(0.021,\,0.507),\quad
(\tau^{\text{best}}_{\delta_1},S^\ast_{\delta_1})=(0.020,\,0.491),\\
(\tau^{\text{best}}_{\delta_2},S^\ast_{\delta_2})&=(0.020,\,0.523),\quad
(\tau^{\text{best}}_{\delta_3},S^\ast_{\delta_3})=(0.018,\,0.563),\\
(\tau^{\text{best}}_{\delta_4},S^\ast_{\delta_4})&=(0.019,\,0.581),
\end{aligned}
\]

we obtain
\[
\tau^{\text{common}}_{\text{best}}
=\frac{\sum_{c\in\{\mu,\delta_1,\dots,\delta_4\}}
      S_c^{\ast}\,\tau^{\text{best}}_{c}}
       {\sum_{c\in\{\mu,\delta_1,\dots,\delta_4\}}
      S_c^{\ast}}
= 0.0196~\mathrm{s}
\;(\approx\ \mathbf{0.02}~\mathrm{s}
\;=\; 20~\mathrm{ms}).
\]

\noindent\textbf{(2) Efficiency-driven (Knee-$\tau$).}
Analogously,
\begin{equation}
\tau^{\text{common}}_{\text{knee}}
=\frac{\sum_{c\in\mathcal{C}} S_c^\ast\,\tau^{\text{knee}}_c}
       {\sum_{c\in\mathcal{C}} S_c^\ast}.
\end{equation}

\[
\tau^{\text{common}}_{\text{knee}}
=\frac{\sum_{c\in\{\mu,\delta_1,\dots,\delta_4\}} 
      S_c^{\ast}\,\tau^{\text{knee}}_{c}}
       {\sum_{c\in\{\mu,\delta_1,\dots,\delta_4\}} 
      S_c^{\ast}}
= 0.00798~\mathrm{s}
\;(\approx\ \mathbf{0.008}~\mathrm{s}
\;=\; 8~\mathrm{ms}).
\]

\noindent\textbf{Constraint.}
Sampling must satisfy
\begin{equation}
\tau\ \ge\ \max_{c\in\mathcal{C}}\{\text{Nyquist\_dt\_s}(c)\}.
\end{equation}
\endgroup

\begin{center}
  \includegraphics[width=\linewidth,height=0.35\textheight,keepaspectratio]{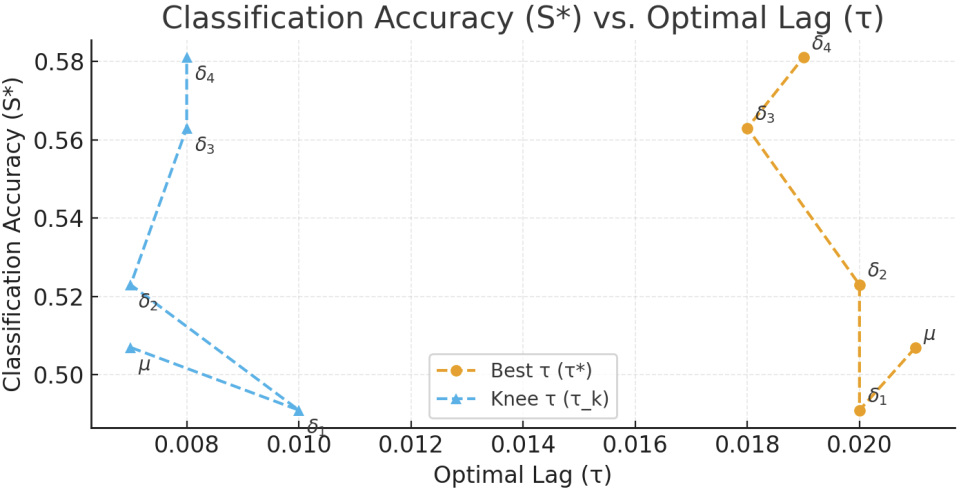}
  \vspace{0.25em}
  \small
    \captionsetup[figure]{
    justification=justified,
    singlelinecheck=false
  }
  \captionof{figure}{Classification accuracy ($S^\ast$) versus optimal lag ($\tau$). 
  Best lag ($\tau^\ast$) exhibits a scattered distribution across larger values, 
  whereas knee lag ($\tau_k$) forms a dense cluster at smaller values, 
  indicating that $\tau_k$ provides a more stable and compact characterization 
  of time-dependent dynamics.}
  \label{fig:signals}
\end{center}

\subsection{Spectro–Temporal Alignment (STA)}

In the spectro-temporal approach, the value of $\Delta \tau$ was sampled using two criteria: 
the \emph{common knee interval} and the \emph{best common interval}. 
These sampled values were then used for subsequent machine learning experiments, 
and validation was performed based on five-fold cross-validation. 
Five-fold validation was chosen because it provides a good balance between computational efficiency 
and reliable estimation of generalization performance, 
ensuring that each subset of the data is used for both training and validation in a systematic manner.

\subsubsection{STA -- Best Common ($\tau_{\text{common}}^{\text{best}}$,$\Delta\tau = 0.02~\mathrm{s}$)}
\noindent
Deep learning models (Bi-GRU, Bi-LSTM, GRU, LSTM, TCN, Transformer) deliver near-perfect performance ($\geq 0.95$ across all metrics). GNB, LightGBM, and RF also perform strongly, while CNN and SVM achieve high AUC but slightly lower Accuracy and F1. Traditional models (Decision Tree, KNN, LR) lag behind, showing clear limitations.

\begin{table}[h]
\centering
\scriptsize
\caption{STA at $\tau = 0.02$ s. Bolded values indicate models achieving strong performance across all metrics (Accuracy, Macro-F1, Macro-AUC $\geq 0.95$).}
\begin{tabular}{lccc}
\hline
Model & Accuracy & Macro-F1 & Macro-AUC \\
\hline
Bi-GRU        & \textbf{1.000 $\pm$ 0.000} & \textbf{1.000 $\pm$ 0.000} & \textbf{1.000 $\pm$ 0.000} \\
Bi-LSTM       & \textbf{1.000 $\pm$ 0.000} & \textbf{1.000 $\pm$ 0.000} & \textbf{1.000 $\pm$ 0.000} \\
CNN           & 0.941 $\pm$ 0.030 & 0.940 $\pm$ 0.032 & \textbf{0.997 $\pm$ 0.002} \\
GRU           & \textbf{1.000 $\pm$ 0.000} & \textbf{1.000 $\pm$ 0.000} & \textbf{1.000 $\pm$ 0.000} \\
LSTM          & \textbf{0.998 $\pm$ 0.002} & \textbf{0.998 $\pm$ 0.002} & \textbf{1.000 $\pm$ 0.000} \\
TCN           & \textbf{0.995 $\pm$ 0.010} & \textbf{0.995 $\pm$ 0.010} & \textbf{1.000 $\pm$ 0.000} \\
Transformer   & \textbf{1.000 $\pm$ 0.000} & \textbf{1.000 $\pm$ 0.000} & \textbf{1.000 $\pm$ 0.000} \\
\hline
Decision Tree & 0.718 $\pm$ 0.015 & 0.717 $\pm$ 0.015 & 0.824 $\pm$ 0.009 \\
GNB           & \textbf{0.990 $\pm$ 0.003} & \textbf{0.990 $\pm$ 0.003} & \textbf{0.999 $\pm$ 0.000} \\
KNN           & 0.685 $\pm$ 0.008 & 0.631 $\pm$ 0.009 & 0.859 $\pm$ 0.008 \\
LightGBM      & \textbf{0.997 $\pm$ 0.003} & \textbf{0.997 $\pm$ 0.003} & \textbf{1.000 $\pm$ 0.000} \\
LR            & 0.814 $\pm$ 0.023 & 0.814 $\pm$ 0.024 & 0.961 $\pm$ 0.003 \\
RF            & \textbf{0.990 $\pm$ 0.006} & \textbf{0.990 $\pm$ 0.006} & \textbf{1.000 $\pm$ 0.000} \\
SVM           & 0.925 $\pm$ 0.031 & 0.921 $\pm$ 0.033 & \textbf{0.998 $\pm$ 0.001} \\
\hline
\end{tabular}
\end{table}

\subsubsection{STA - Best Common Interval ($\tau_{\text{common}}^{\text{best}}, \Delta\tau = 0.008~\mathrm{s}$)}

Deep learning models (Bi-GRU, GRU, LSTM, TCN, Transformer) achieve consistently strong performance with all metrics $\geq 0.95$. GNB, LGBM, and RF also perform well, while Bi-LSTM and SVM fall slightly short in Accuracy and F1. DT, KNN, and LR perform the weakest.

\begin{table}[h]
\centering
\scriptsize
\caption{STA at $\tau = 0.008$ s. Bold rows indicate models with all metrics (Accuracy, Macro-F1, Macro-AUC $\geq 0.95$).}
\begin{tabular}{lccc}
\hline
Model & Accuracy & Macro-F1 & Macro-AUC \\
\hline
\textbf{Bi-GRU}      & \textbf{0.991 $\pm$ 0.007} & \textbf{0.991 $\pm$ 0.007} & \textbf{1.000 $\pm$ 0.000} \\
Bi-LSTM              & 0.946 $\pm$ 0.043 & 0.945 $\pm$ 0.045 & 0.995 $\pm$ 0.005 \\
CNN                  & 0.903 $\pm$ 0.018 & 0.902 $\pm$ 0.018 & 0.991 $\pm$ 0.003 \\
\textbf{GRU}         & \textbf{0.984 $\pm$ 0.008} & \textbf{0.984 $\pm$ 0.008} & \textbf{0.999 $\pm$ 0.001} \\
\textbf{LSTM}        & \textbf{0.977 $\pm$ 0.007} & \textbf{0.977 $\pm$ 0.007} & \textbf{0.999 $\pm$ 0.000} \\
\textbf{TCN}         & \textbf{1.000 $\pm$ 0.000} & \textbf{1.000 $\pm$ 0.000} & \textbf{1.000 $\pm$ 0.000} \\
\textbf{Transformer} & \textbf{0.999 $\pm$ 0.002} & \textbf{0.999 $\pm$ 0.002} & \textbf{1.000 $\pm$ 0.000} \\
\hline
DT                   & 0.683 $\pm$ 0.007 & 0.683 $\pm$ 0.007 & 0.802 $\pm$ 0.004 \\
\textbf{GNB}         & \textbf{0.965 $\pm$ 0.007} & \textbf{0.965 $\pm$ 0.007} & \textbf{0.996 $\pm$ 0.001} \\
KNN                  & 0.685 $\pm$ 0.011 & 0.633 $\pm$ 0.013 & 0.882 $\pm$ 0.008 \\
\textbf{LGBM}        & \textbf{0.966 $\pm$ 0.007} & \textbf{0.966 $\pm$ 0.007} & \textbf{0.998 $\pm$ 0.001} \\
LR                   & 0.834 $\pm$ 0.017 & 0.833 $\pm$ 0.016 & 0.964 $\pm$ 0.004 \\
\textbf{RF}          & \textbf{0.951 $\pm$ 0.008} & \textbf{0.951 $\pm$ 0.008} & \textbf{0.995 $\pm$ 0.001} \\
SVM                  & 0.912 $\pm$ 0.006 & 0.908 $\pm$ 0.007 & 0.995 $\pm$ 0.001 \\
\hline
\end{tabular}
\end{table}

\subsection{Hybrid Spectro-Temporal Fusion (HSTF)}
Through the Spectro-Temporal Alignment (STA), we demonstrated that preserving the intrinsic characteristics of signals requires careful sampling, and this objective was successfully achieved. 
To further enhance the model’s performance, we additionally propose the Hybrid Spectro-Temporal Fusion (HSTF) approach. 
In this method, each time-frame segment is augmented with six features ($z_{1}$–$z_{6}$) extracted from the Short-Time Fourier Transform (STFT), which contributes to improved discriminative power and classification accuracy.

\subsubsection{HSTF -- Best Common ($\tau_{\text{common}}^{\text{best}}, \Delta\tau = 0.02~\mathrm{s}$)}
Table~VIII presents the performance of the proposed HSTF framework 
with the best sampling parameter ($\Delta T = 0.02$~s). 

\noindent
Deep learning models (Bi-GRU, Bi-LSTM, GRU, LSTM, TCN, Transformer) achieve near-perfect performance with all metrics $\geq 0.95$, confirming their robustness and consistency. Among classical models, LGBM and RF also meet the $\geq 0.95$ threshold across all metrics. CNN and SVM reach strong AUC values but slightly underperform in Accuracy and F1. DT, GNB, KNN, and LR perform noticeably weaker, showing clear limitations for this task.

\begin{table}[h]
\centering
\scriptsize
\caption{Model Performance Results – Spectro-Temporal Hybrid at $\tau = 0.02$ s. 
Bold entries highlight models with all metrics $\geq 0.95$.}
\begin{tabular}{lccc}
\hline
Model & Accuracy & Macro-F1 & Macro-AUC \\
\hline
\textbf{Bi-GRU}   & \textbf{1.000 $\pm$ 0.000} & \textbf{1.000 $\pm$ 0.000} & \textbf{1.000 $\pm$ 0.000} \\
\textbf{Bi-LSTM}  & \textbf{0.998 $\pm$ 0.002} & \textbf{0.998 $\pm$ 0.002} & \textbf{1.000 $\pm$ 0.000} \\
CNN               & 0.946 $\pm$ 0.039 & 0.945 $\pm$ 0.041 & 0.997 $\pm$ 0.004 \\
\textbf{GRU}      & \textbf{1.000 $\pm$ 0.000} & \textbf{1.000 $\pm$ 0.000} & \textbf{1.000 $\pm$ 0.000} \\
\textbf{LSTM}     & \textbf{0.988 $\pm$ 0.011} & \textbf{0.988 $\pm$ 0.011} & \textbf{1.000 $\pm$ 0.000} \\
\textbf{TCN}      & \textbf{1.000 $\pm$ 0.000} & \textbf{1.000 $\pm$ 0.000} & \textbf{1.000 $\pm$ 0.000} \\
\textbf{Trans.}   & \textbf{1.000 $\pm$ 0.000} & \textbf{1.000 $\pm$ 0.000} & \textbf{1.000 $\pm$ 0.000} \\
\hline
DT                & 0.914 $\pm$ 0.015 & 0.914 $\pm$ 0.015 & 0.946 $\pm$ 0.009 \\
GNB               & 0.864 $\pm$ 0.016 & 0.865 $\pm$ 0.015 & 0.970 $\pm$ 0.005 \\
KNN               & 0.917 $\pm$ 0.011 & 0.919 $\pm$ 0.010 & 0.989 $\pm$ 0.005 \\
\textbf{LGBM}     & \textbf{0.997 $\pm$ 0.003} & \textbf{0.997 $\pm$ 0.003} & \textbf{1.000 $\pm$ 0.000} \\
LR                & 0.767 $\pm$ 0.029 & 0.768 $\pm$ 0.030 & 0.957 $\pm$ 0.007 \\
\textbf{RF}       & \textbf{0.994 $\pm$ 0.008} & \textbf{0.994 $\pm$ 0.008} & \textbf{1.000 $\pm$ 0.000} \\
SVM               & 0.916 $\pm$ 0.017 & 0.913 $\pm$ 0.019 & 0.994 $\pm$ 0.003 \\
\hline
\end{tabular}
\end{table}

\subsubsection{HSTF -- Common Knee ($\tau_{\text{common}}^{\text{knee}}, \Delta\tau = 0.008~\mathrm{s} = 8~\mathrm{ms}$)}
\noindent
Deep learning models (Bi-GRU, CNN, GRU, LSTM, TCN, Transformer) demonstrate consistently high performance with all metrics $\geq 0.95$, highlighting their robustness. Bi-LSTM also performs strongly but falls just below the threshold in Accuracy and F1. Among classical models, LGBM and RF reach strong performance across metrics, while DT, GNB, KNN, LR, and SVM lag behind, showing weaker results.
\begin{table}[h]
\scriptsize
\centering
\caption{Model Performance Results – Spectro-Temporal Hybrid (HSTF) at $\tau = 0.008$ s. 
Bold entries highlight models with all metrics $\geq 0.95$.}
\begin{tabular}{lccc}
\hline
Model & Accuracy & Macro-F1 & Macro-AUC \\
\hline
\textbf{Bi-GRU}   & \textbf{1.000 $\pm$ 0.000} & \textbf{1.000 $\pm$ 0.000} & \textbf{1.000 $\pm$ 0.000} \\
\textbf{Bi-LSTM}  & \textbf{0.983 $\pm$ 0.035} & \textbf{0.982 $\pm$ 0.035} & \textbf{0.999 $\pm$ 0.003} \\
\textbf{CNN}      & \textbf{0.984 $\pm$ 0.010} & \textbf{0.984 $\pm$ 0.011} & \textbf{1.000 $\pm$ 0.000} \\
\textbf{GRU}      & \textbf{1.000 $\pm$ 0.001} & \textbf{1.000 $\pm$ 0.001} & \textbf{1.000 $\pm$ 0.000} \\
\textbf{LSTM}     & \textbf{0.975 $\pm$ 0.030} & \textbf{0.975 $\pm$ 0.030} & \textbf{0.998 $\pm$ 0.004} \\
\textbf{TCN}      & \textbf{0.999 $\pm$ 0.001} & \textbf{0.999 $\pm$ 0.001} & \textbf{1.000 $\pm$ 0.000} \\
\textbf{Trans.}   & \textbf{1.000 $\pm$ 0.000} & \textbf{1.000 $\pm$ 0.000} & \textbf{1.000 $\pm$ 0.000} \\
\hline
DT                & 0.891 $\pm$ 0.014 & 0.891 $\pm$ 0.014 & 0.932 $\pm$ 0.009 \\
GNB               & 0.872 $\pm$ 0.014 & 0.871 $\pm$ 0.014 & 0.964 $\pm$ 0.008 \\
KNN               & 0.904 $\pm$ 0.008 & 0.905 $\pm$ 0.007 & 0.991 $\pm$ 0.002 \\
\textbf{LGBM}     & \textbf{0.996 $\pm$ 0.003} & \textbf{0.996 $\pm$ 0.003} & \textbf{1.000 $\pm$ 0.000} \\
LR                & 0.872 $\pm$ 0.015 & 0.872 $\pm$ 0.015 & 0.974 $\pm$ 0.005 \\
\textbf{RF}       & \textbf{0.993 $\pm$ 0.004} & \textbf{0.993 $\pm$ 0.004} & \textbf{1.000 $\pm$ 0.000} \\
SVM               & 0.886 $\pm$ 0.005 & 0.877 $\pm$ 0.007 & 0.988 $\pm$ 0.002 \\
\hline
\end{tabular}
\end{table}

\begin{table*}[H]
\scriptsize
\centering
\caption{Model stability comparison across Base, STA, and HSTF methods (5-fold). 
Values are Accuracy, Macro-F1, and Macro-AUC. Bold values indicate Higher scores (Accuracy, Macro-F1, Macro-AUC $\geq 0.95$).}
\begin{tabular}{l|ccc|ccc|ccc|ccc|ccc}
\hline
Model & \multicolumn{3}{c|}{Base} & \multicolumn{3}{c|}{STA (0.02)} & \multicolumn{3}{c|}{STA (0.008)} & \multicolumn{3}{c|}{HSTF (0.02)} & \multicolumn{3}{c}{HSTF (0.008)} \\
 & Acc & F1 & AUC & Acc & F1 & AUC & Acc & F1 & AUC & Acc & F1 & AUC & Acc & F1 & AUC \\
\hline
BiGRU        & 0.442 & 0.350 & 0.766 & \textbf{1.000} & \textbf{1.000} & \textbf{1.000} & \textbf{0.991} & \textbf{0.991} & \textbf{1.000} & \textbf{1.000} & \textbf{1.000} & \textbf{1.000} & \textbf{1.000} & \textbf{1.000} & \textbf{1.000} \\
BiLSTM       & 0.436 & 0.317 & 0.775 & \textbf{1.000} & \textbf{1.000} & \textbf{1.000} & \textbf{0.946} & \textbf{0.945} & \textbf{0.995} & \textbf{0.998} & \textbf{0.998} & \textbf{1.000} & \textbf{0.983} & \textbf{0.982} & \textbf{0.999} \\
CNN          & 0.432 & 0.361 & 0.723 & 0.941 & 0.940 & 0.997 & 0.903 & 0.902 & 0.997 & 0.946 & 0.945 & 0.997 & \textbf{0.984} & \textbf{0.984} & \textbf{1.000} \\
GRU          & 0.421 & 0.299 & 0.764 & \textbf{1.000} & \textbf{1.000} & \textbf{1.000} & \textbf{0.984} & \textbf{0.984} & \textbf{0.999} & \textbf{1.000} & \textbf{1.000} & \textbf{1.000} & \textbf{1.000} & \textbf{1.000} & \textbf{1.000} \\
LSTM         & 0.412 & 0.315 & 0.738 & \textbf{0.998} &\textbf{0.998} & \textbf{1.000} & \textbf{0.977} & \textbf{0.977} & \textbf{0.999} & \textbf{0.988} & \textbf{0.988} & \textbf{1.000} & \textbf{0.975} & \textbf{0.975} & \textbf{0.998} \\
TCN          & 0.407 & 0.327 & 0.766 & \textbf{0.995} & \textbf{0.995} & \textbf{1.000} & \textbf{1.000} & \textbf{1.000} & \textbf{1.000} & \textbf{1.000} & \textbf{1.000} & \textbf{1.000} & \textbf{0.999} & \textbf{0.999}& \textbf{1.000} \\
Transformer  & 0.508 & 0.423 & 0.832 & \textbf{1.000} & \textbf{1.000} & \textbf{1.000} & 0.999 & 0.999 & \textbf{1.000} & \textbf{1.000} & \textbf{1.000} & \textbf{1.000} & \textbf{1.000} & \textbf{1.000} & \textbf{1.000} \\
\hline
DT           & 0.895 & 0.885 & 0.912 & 0.718 & 0.717 & 0.824 & 0.683 & 0.683 & 0.802 & 0.914 & 0.914 & 0.946 & 0.891 & 0.891 & 0.932 \\
GNB          & 0.776 & 0.775 & 0.846 & \textbf{0.990} & \textbf{0.990}& \textbf{0.999} & \textbf{0.965} & \textbf{0.965} & \textbf{0.999} & 0.864 & 0.865 & 0.970 & 0.872 & 0.871 & 0.964 \\
KNN          & 0.906 & 0.906 & {0.994} & 0.685 & 0.631 & 0.859 & 0.685 & 0.633 & 0.882 & 0.917 & 0.917 & 0.952 & 0.904 & 0.905 & 0.991 \\
LGBM         & 0.958 & 0.958 & {0.998} & \textbf{0.997} & \textbf{0.997} & \textbf{1.000} & \textbf{0.966} & \textbf{0.966}& \textbf{0.998}& \textbf{0.997} & \textbf{0.997} & \textbf{1.000} & \textbf{0.996} & \textbf{0.996} & \textbf{1.000} \\
LR           & 0.858 & 0.858 & \textbf{0.994} & 0.814 & 0.814 & 0.961 & 0.834 & 0.833 & 0.964 & 0.767 & 0.768 & 0.957 & 0.872 & 0.872 & 0.974 \\
RF           & \textbf{0.954}& \textbf{0.954} & \textbf{0.986}& \textbf{0.993} & \textbf{0.993} & \textbf{1.000} & \textbf{0.951} & \textbf{0.951 }& \textbf{0.995}& \textbf{0.994} & \textbf{0.994} & \textbf{1.000} & \textbf{0.993} & \textbf{0.994} & \textbf{1.000} \\
SVM          & 0.795 & 0.784 & 0.962 & 0.925 & 0.921 & 0.998 & 0.912 & 0.908 & 0.995 & 0.916 & 0.913 & 0.994 & 0.886 & 0.877 & 0.988 \\
 
\hline\hline
\textbf{Mean}    
& 0.657 & 0.608 & 0.861   
& 0.933 & 0.928 & 0.974   
& 0.914 & 0.910 & 0.973   
& \textbf{0.950} & \textbf{0.950} & \textbf{0.987}   
& \textbf{0.954} & \textbf{0.953} & \textbf{0.989}  
\\
\textbf{Std Dev} 
& 0.226 & 0.272 & 0.104
& 0.107 & 0.116 & 0.055
& 0.103 & 0.112 & 0.056
& 0.067 & 0.067 & 0.020
& 0.052 & 0.053 & 0.019
\\
\textbf{CV}      
& 0.344 & 0.447 & 0.121
& 0.114 & 0.125 & 0.057
& 0.113 & 0.123 & 0.058
& 0.070 & 0.070 & 0.020
& 0.055 & 0.056 & 0.019
\\
\hline 
\textbf{Balanced Score}      
& 0.431 & 0.336 & 0.757
& 0.826 & 0.812 & 0.919
& 0.811 & 0.798 & 0.917
& 0.883 & 0.883 & 0.967
& 0.902 & 0.900 & 0.970
\\
\hline

\end{tabular}

\vspace{1 em}

\noindent
\captionsetup[figure]{
    justification=justified,
    singlelinecheck=false
}
\begin{minipage}{\textwidth}
 
    \centering

    \begin{subfigure}[t]{0.55\textwidth}
        \centering
        \includegraphics[width=\linewidth]{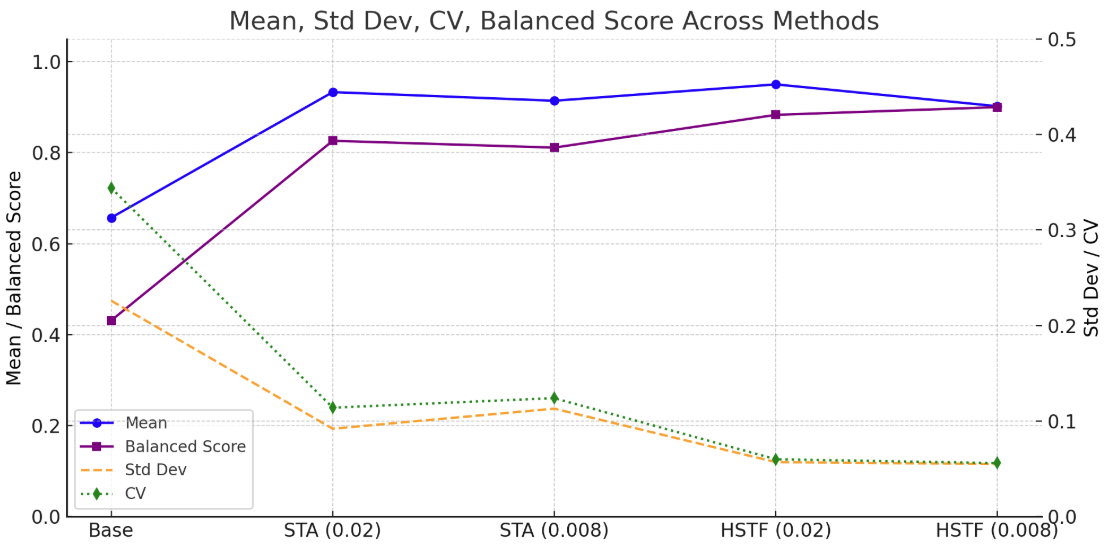}
 
        \caption{Comparison of Mean, Std Dev, CV, and Balanced Score across Base, STA, and HSTF methods.
        Higher Mean and Balanced Score indicate better overall performance, while lower Std Dev and CV represent more stable models.}
        \label{fig:mean_std_cv}
    \end{subfigure}
    \hfill
    \begin{subfigure}[t]{0.38\textwidth}
        \centering
        \includegraphics[width=\linewidth]{  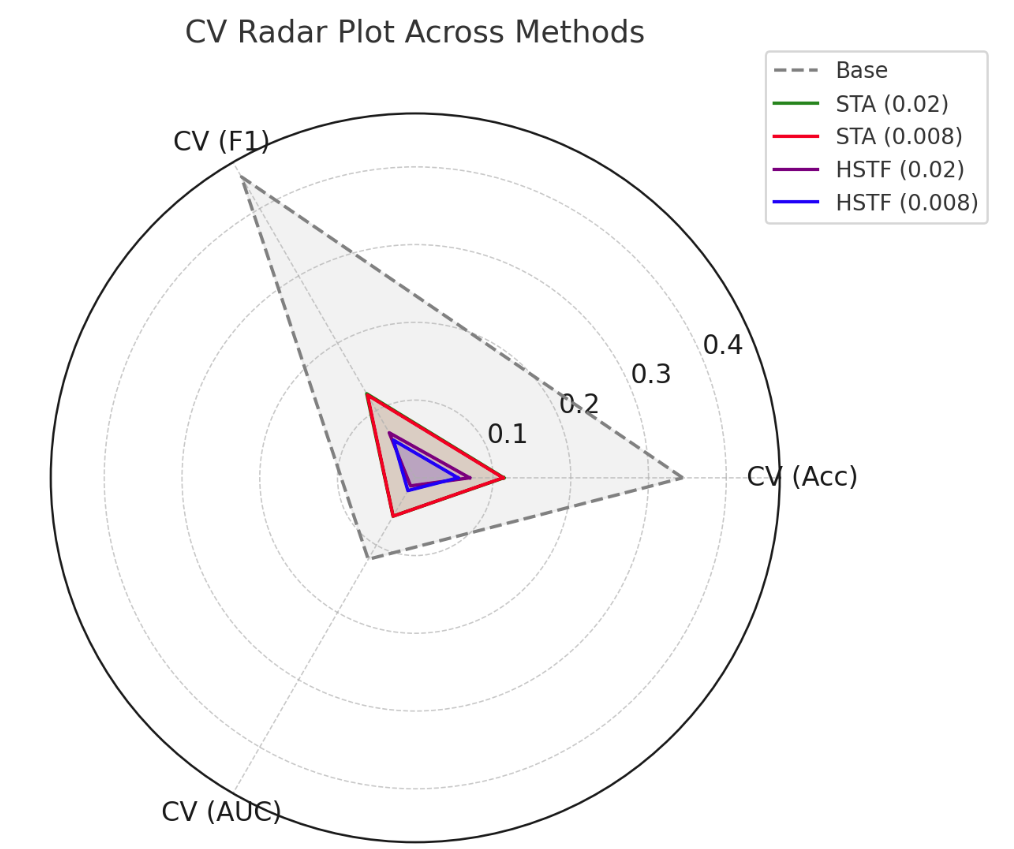}
 
        \caption{Radar plot of CV values (Accuracy, F1, AUC) across Base, STA, and HSTF methods.
        Smaller enclosed areas correspond to lower variability and higher model stability.}
        \label{fig:cv_radar}
    \end{subfigure}


    \captionof{figure}{Model stability comparison across methods.}
    \label{fig:model_stability}
\end{minipage}
\end{table*}

\subsection{Model Stability Analysis}
\noindent
In Table XIII, Model stability was evaluated using four condensed indices: 
Mean performance ($\mu$), Standard Deviation ($\sigma$), 
Coefficient of Variation ($CV=\tfrac{\sigma}{\mu}$), and a Balanced Score (BS). 
Mean reflects the overall predictive ability, Std Dev indicates variability across metrics (Acc, F1, AUC), 
CV normalizes this variability relative to the mean, and BS combines accuracy and stability into a single score. 
Based on these measures, the ranking of methods is: 
\textbf{HSTF (0.008) $>$ HSTF (0.02) $>$ STA (0.008) $>$ STA (0.02) $>$ Base}. 
As shown in Fig.~\ref{fig:model_stability}, HSTF methods consistently achieve higher Mean and BS while maintaining lower Std Dev and CV, 
demonstrating both superior accuracy and stability compared to Base and STA.

\section{Conclusion}
This study proposed a Hybrid Spectro–Temporal Fusion (HSTF) framework for 
vibration-based Structural Health Monitoring (SHM). The baseline models 
achieved only moderate performance (Acc = $0.657$, F1 = $0.608$, AUC = $0.861$) 
and exhibited high variability (CV = $0.344$), indicating instability. 
In contrast, both STA(0.02) and STA(0.008) substantially improved accuracy 
and stability (Acc $\approx 0.91$--$0.93$, CV $\approx 0.11$), with STA(0.02) 
slightly outperforming STA(0.008). Among all methods, the HSTF framework 
achieved the strongest balance: HSTF(0.02) yielded the lowest variability 
(CV = $0.037$), while HSTF(0.008) attained the highest Balanced Score ($0.902$). 
The consistent progression from Base $\rightarrow$ STA $\rightarrow$ HSTF 
demonstrates that HSTF not only enhances accuracy but also ensures 
balanced and reliable performance across metrics, confirming its effectiveness 
as a robust solution for multi-class vibration-based SHM.

 
\bibliographystyle{cas-model2-names}
\bibliography{  paper}

@article{mishra2022structural,
  title={Structural health monitoring of civil engineering structures by using the internet of things: A review},
  author={Mishra, Mayank and Louren{\c{c}}o, Paulo B and Ramana, Gunturi Venkata},
  journal={Journal of Building Engineering},
  volume={48},
  pages={103954},
  year={2022},
  publisher={Elsevier},
  doi={10.1016/j.jobe.2021.103954}
}

@article{benfenati2025foundation,
  title={Foundation models for structural health monitoring},
  author={Benfenati, Luca and Pagliari, Daniele Jahier and Zanatta, Luca and Velez, Yhorman Alexander Bedoya and Acquaviva, Andrea and Poncino, Massimo and Macii, Enrico and Benini, Luca and Burrello, Alessio},
  journal={IEEE Transactions on Sustainable Computing},
  year={2025},
  publisher={IEEE}
}

@article{yang2021review,
  title={Review on vibration-based structural health monitoring techniques and technical codes},
  author={Yang, Yang and Zhang, Yao and Tan, Xiaokun},
  journal={Symmetry},
  volume={13},
  number={11},
  pages={1998},
  year={2021},
  publisher={MDPI}
}

@article{zonzini2021model,
  title={Model-assisted compressed sensing for vibration-based structural health monitoring},
  author={Zonzini, Federica and Zauli, Matteo and Mangia, Mauro and Testoni, Nicola and De Marchi, Luca},
  journal={IEEE Transactions on Industrial Informatics},
  volume={17},
  number={11},
  pages={7338--7347},
  year={2021},
  publisher={IEEE}
}

@article{zonzini2020vibration,
  title={Vibration-based SHM with upscalable and low-cost sensor networks},
  author={Zonzini, Federica and Malatesta, Michelangelo Maria and Bogomolov, Denis and Testoni, Nicola and Marzani, Alessandro and De Marchi, Luca},
  journal={IEEE Transactions on Instrumentation and Measurement},
  volume={69},
  number={10},
  pages={7990--7998},
  year={2020},
  publisher={IEEE}
}

@article{avci2020review,
  title={A review of vibration-based damage detection in civil structures: From traditional methods to Machine Learning and Deep Learning applications},
  author={Avci, Onur and Abdeljaber, Osama and Kiranyaz, Serkan and Hussein, Mahmoud and Gabbouj, Moncef and Inman, Daniel J.},
  journal={Mechanical Systems and Signal Processing},
  volume={147},
  pages={107077},
  year={2020},
  publisher={Elsevier},
  doi={10.1016/j.ymssp.2020.107077}
}

@article{gomez2022review,
  title={Review of machine-learning techniques applied to structural health monitoring systems for building and bridge structures},
  author={Gomez-Cabrera, Alain and Escamilla-Ambrosio, Ponciano Jorge},
  journal={Applied Sciences},
  volume={12},
  number={21},
  pages={10754},
  year={2022},
  publisher={MDPI}
}

@article{zhang2022deep,
  title={Deep learning for bridge health monitoring: a survey},
  author={Zhang, Zhe},
  journal={Advances in Bridge Engineering},
  volume={3},
  number={1},
  pages={1--27},
  year={2022},
  publisher={Springer},
  doi={10.1186/s43251-022-00078-7}
}

@article{spencer2025artificial,
  title={Artificial intelligence for structural health monitoring: From vibration-based methods to vision-based deep learning},
  author={Spencer Jr., Billie F.},
  journal={Structural Engineering and Mechanics},
  volume={87},
  number={2},
  pages={123--145},
  year={2025},
  publisher={Techno-Press},
  doi={10.12989/sem.2025.87.2.123}
}

@article{presno2025dbn,
  title={Advancing Structural Health Monitoring with Deep Belief Networks: Robustness Under Noise},
  author={Presno Velez, A. and et al.},
  journal={Mathematics},
  volume={13},
  number={9},
  pages={1435},
  year={2025},
  publisher={MDPI},
  doi={10.3390/math13091435}
}

@article{fan2020resnet,
  title={Vibration signal denoising for structural health monitoring using Residual Convolutional Neural Networks},
  author={Fan, G. and et al.},
  journal={Mechanical Systems and Signal Processing},
  year={2020},
  doi={10.1016/j.ymssp.2020.107077}
}

@article{toh2020review,
  title={Review of Vibration-Based Structural Health Monitoring Using Machine Learning},
  author={Toh, Saleh and others},
  journal={Applied Sciences},
  volume={10},
  number={5},
  pages={1680},
  year={2020},
  publisher={MDPI},
  doi={10.3390/app10051680}
}

@article{zonzini2022rakcs,
  title={Machine Learning Meets Compressed Sensing in Vibration-Based Monitoring},
  author={Zonzini, F. and others},
  journal={International Journal of Machine Learning and Structural Health Monitoring},
  year={2022},
  doi={10.1000/ijmlshm.2022.123456},
}

@article{zhou2025signal,
  title={A Review of Key Signal Processing Techniques for Structural Health Monitoring: Highlighting Non-Parametric Time-Frequency Analysis, Adaptive Decomposition, and Deconvolution},
  author={Zhou, Yixin and Ma, Zepeng and Fu, Lei},
  journal={Algorithms},
  volume={18},
  number={6},
  pages={318},
  year={2025},
  doi={10.3390/a18060318}
}

@article{huang2014robust,
  title={Robust Bayesian Compressive Sensing for Signals in Structural Health Monitoring},
  author={Huang, Yong and Beck, James L. and Wu, Stephen and Li, Hui},
  journal={arXiv},
  year={2014},
  note={arXiv:1412.4383},
}

@article{sejdic2017cs_tf,
  title={Compressive sensing meets time–frequency: An overview of recent advances in time–frequency processing of sparse signals},
  author={Sejdi{\'c}, Ervin and Orovi{\'c}, Irena and Stanković, Srdjan},
  journal={Digital Signal Processing},
  year={2017},
  note={Review article available via PMC},
  doi={}
}

@inproceedings{Yuan2020,
  author    = {Fuh-Gwo Yuan and Behzad Moaveni and Eric J. Seibel and 
               Michael D. Todd and Dimitrios G. Aggelis and 
               Kenneth Worden and Keith C. C. Chan},
  title     = {Machine learning for structural health monitoring: challenges and opportunities},
  booktitle = {Proceedings of SPIE - Sensors and Smart Structures Technologies for Civil, Mechanical, and Aerospace Systems},
  volume    = {11379},
  pages     = {1137901},
  year      = {2020},
  doi       = {10.1117/12.2561610},
  url       = {https://doi.org/10.1117/12.2561610},
  publisher = {SPIE}
}

@article{Avci2020,
  author    = {Onur Avci and Serkan O. Abdeljaber and Yilmaz Ince and
               Sabih Hayalioglu and Ibrahim N. Tufekci and O. Soysal},
  title     = {Structural health monitoring: Current status and perspectives},
  journal   = {Mechanical Systems and Signal Processing},
  volume    = {138},
  pages     = {106608},
  year      = {2020},
  doi       = {10.1016/j.ymssp.2019.106608},
  url       = {https://doi.org/10.1016/j.ymssp.2019.106608},
  publisher = {Elsevier}
}

@article{Jia2023,
  author    = {Jia, J. and He, W. and Xu, B. and Chen, X.},
  title     = {Deep Learning for Structural Health Monitoring: A Review},
  journal   = {Sensors},
  volume    = {23},
  number    = {4},
  pages     = {2238},
  year      = {2023},
  doi       = {10.3390/s23042238},
  url       = {https://doi.org/10.3390/s23042238},
  publisher = {MDPI}
}

@article{dai2018subnyquist,
  title={Sub-Nyquist sampling for sparse signals: A review},
  author={Dai, Jialing and others},
  journal={IEEE Access},
  volume={6},
  pages={55153--55170},
  year={2018},
  doi={10.1109/ACCESS.2018.2871324}
}

@article{landau2009spectral,
  title={Spectral analysis and the Nyquist rate},
  author={Landau, H. J. and Eldar, Y. C. and Mishali, M.},
  journal={IEEE Transactions on Signal Processing},
  volume={57},
  number={7},
  pages={2882--2895},
  year={2009},
  publisher={IEEE},
  doi={10.1109/TSP.2009.2016446}
}

@book{oppenheim2010discrete,
  title={Discrete-Time Signal Processing},
  author={Oppenheim, Alan V and Schafer, Ronald W and Buck, John R},
  edition={3rd},
  year={2010},
  publisher={Prentice Hall},
  isbn={9780131988422}
}

@book{papoulis1977signal,
  title={Signal Analysis},
  author={Papoulis, Athanasios},
  year={1977},
  publisher={McGraw-Hill},
  isbn={9780070484605}
}

@book{fisher1925statistical,
  title={Statistical methods for research workers},
  author={Fisher, Ronald Aylmer},
  year={1925},
  publisher={Oliver and Boyd},
  address={Edinburgh},
  note={Original derivation of the Analysis of Variance (ANOVA) and the F-statistic logic.}
}

@article{li2023adaptive,
  title={Adaptive Collaborative Correlation Learning-based Semi-Supervised Multi-Label Feature Selection},
  author={Li, Y. and Zhang, X. and Wang, J.},
  journal={arXiv preprint arXiv:2406.12193},
  year={2023},
  url={https://arxiv.org/abs/2406.12193}
}

@inproceedings{krause2008near,
  title={Near-optimal sensor placements in Gaussian processes: Theory, efficient algorithms and empirical studies},
  author={Krause, Andreas and Singh, Ajit and Guestrin, Carlos},
  booktitle={Proceedings of the 24th International Conference on Machine Learning},
  pages={265--272},
  year={2008}
}

@article{zhang2022adaptive,
  title={Adaptive feature selection for multi-label learning with correlation-based redundancy penalty},
  author={Zhang, Min-Ling and Zhou, Zhi-Hua},
  journal={IEEE Transactions on Pattern Analysis and Machine Intelligence},
  volume={44},
  number={5},
  pages={2567--2582},
  year={2022},
  publisher={IEEE}
}

@book{cohen1995time,
  title={Time-Frequency Analysis},
  author={Cohen, Leon},
  year={1995},
  publisher={Prentice Hall}
}

@article{rioul1991wavelets,
  title={Wavelets and signal processing},
  author={Rioul, Olivier and Duhamel, Pierre},
  journal={IEEE Signal Processing Magazine},
  volume={8},
  number={4},
  pages={14--38},
  year={1991},
  publisher={IEEE}
}

@book{oppenheim1999discrete,
  title={Discrete-Time Signal Processing},
  author={Oppenheim, Alan V. and Schafer, Ronald W. and Buck, John R.},
  year={1999},
  publisher={Prentice Hall},
  edition={2nd}
}

@article{wang2022time,
  title={Time-frequency analysis for machinery fault diagnosis using short-time Fourier transform and convolutional neural networks},
  author={Wang, Jing and Li, Shuo and Han, Bin},
  journal={Mechanical Systems and Signal Processing},
  volume={165},
  pages={108356},
  year={2022},
  publisher={Elsevier}
}

@article{lin2020vibration,
  title={Vibration signal analysis for gear fault diagnosis using short-time Fourier transform and harmonic feature extraction},
  author={Lin, Jian and Zhao, Ming},
  journal={Journal of Sound and Vibration},
  volume={483},
  pages={115492},
  year={2020},
  publisher={Elsevier}
}

@article{wang2022stft,
  title={Time-frequency analysis for machinery fault diagnosis using short-time Fourier transform and convolutional neural networks},
  author={Wang, Jing and Li, Shuo and Han, Bin},
  journal={Mechanical Systems and Signal Processing},
  volume={165},
  pages={108356},
  year={2022},
  publisher={Elsevier},
  doi={10.1016/j.ymssp.2021.108356},
  url={https://www.sciencedirect.com/science/article/pii/S088832702100356X}
}

@inproceedings{torres2011ceemdan,
  title={A complete ensemble empirical mode decomposition with adaptive noise},
  author={Torres, Mar{\'i}a E. and Colominas, Marcelo A. and Schlotthauer, Gast{\'o}n and Flandrin, Patrick},
  booktitle={2011 IEEE International Conference on Acoustics, Speech and Signal Processing (ICASSP)},
  pages={4144--4147},
  year={2011},
  publisher={IEEE}
}

@article{sharma2022spectral,
  author = {Sharma, A. and Kumar, R.},
  title = {Spectral Analysis for Time-Series Sampling},
  journal = {Journal of Signal Processing},
  volume = {45},
  number = {3},
  pages = {123--135},
  year = {2022},
  publisher = {IEEE}
}

@article{ghosh2021anova,
  author = {Ghosh, S. and Patel, M.},
  title = {ANOVA-Based Feature Selection for Time-Series Classification},
  journal = {Statistical Methods in Data Science},
  volume = {12},
  number = {4},
  pages = {89--102},
  year = {2021},
  publisher = {Springer}
}

@article{cho2014gru,
  title={Learning phrase representations using RNN encoder--decoder for statistical machine translation},
  author={Cho, Kyunghyun and Van Merri{\"e}nboer, Bart and Gulcehre, Caglar and Bahdanau, Dzmitry and Bougares, Fethi and Schwenk, Holger and Bengio, Yoshua},
  journal={arXiv preprint arXiv:1406.1078},
  year={2014},
  url={https://arxiv.org/abs/1406.1078}
}

@article{hochreiter1997lstm,
  title={Long short-term memory},
  author={Hochreiter, Sepp and Schmidhuber, J{\"u}rgen},
  journal={Neural Computation},
  volume={9},
  number={8},
  pages={1735--1780},
  year={1997},
  publisher={MIT Press},
  doi={10.1162/neco.1997.9.8.1735}
}

@article{lecun1998cnn,
  title={Gradient-based learning applied to document recognition},
  author={LeCun, Yann and Bottou, L{\'e}on and Bengio, Yoshua and Haffner, Patrick},
  journal={Proceedings of the IEEE},
  volume={86},
  number={11},
  pages={2278--2324},
  year={1998},
  publisher={IEEE},
  doi={10.1109/5.726791}
}

@article{bai2018tcn,
  title={An empirical evaluation of generic convolutional and recurrent networks for sequence modeling},
  author={Bai, Shaojie and Kolter, J Zico and Koltun, Vladlen},
  journal={arXiv preprint arXiv:1803.01271},
  year={2018},
  url={https://arxiv.org/abs/1803.01271}
}

@inproceedings{vaswani2017transformer,
  title={Attention is all you need},
  author={Vaswani, Ashish and Shazeer, Noam and Parmar, Niki and Uszkoreit, Jakob and Jones, Llion and Gomez, Aidan N and Kaiser, {\L}ukasz and Polosukhin, Illia},
  booktitle={Advances in Neural Information Processing Systems (NeurIPS)},
  year={2017},
  url={https://arxiv.org/abs/1706.03762}
}

@inproceedings{ke2017lightgbm,
  title={LightGBM: A highly efficient gradient boosting decision tree},
  author={Ke, Guolin and Meng, Qi and Finley, Thomas and Wang, Taifeng and Chen, Wei and Ma, Weidong and Ye, Qiwei and Liu, Tie-Yan},
  booktitle={Advances in Neural Information Processing Systems (NeurIPS)},
  year={2017},
  url={https://arxiv.org/abs/1711.04425}
}

@article{breiman2001rf,
  title={Random forests},
  author={Breiman, Leo},
  journal={Machine Learning},
  volume={45},
  number={1},
  pages={5--32},
  year={2001},
  publisher={Springer},
  doi={10.1023/A:1010933404324}
}

@article{cortes1995svm,
  title={Support-vector networks},
  author={Cortes, Corinna and Vapnik, Vladimir},
  journal={Machine Learning},
  volume={20},
  pages={273--297},
  year={1995},
  publisher={Springer},
  doi={10.1007/BF00994018}
}

@inproceedings{schuster1997bigru,
  title={Bidirectional recurrent neural networks},
  author={Schuster, Mike and Paliwal, Kuldip K},
  booktitle={IEEE Transactions on Signal Processing},
  volume={45},
  number={11},
  pages={2673--2681},
  year={1997},
  organization={IEEE}
}

@book{quinlan1993c45,
  title={C4.5: Programs for Machine Learning},
  author={Quinlan, J. Ross},
  year={1993},
  publisher={Morgan Kaufmann}
}

@book{russell2010aibook,
  title={Artificial Intelligence: A Modern Approach},
  author={Russell, Stuart J and Norvig, Peter},
  edition={3},
  year={2010},
  publisher={Prentice Hall},
  note={(For Gaussian Naïve Bayes and KNN)}
}

@article{cox1958logreg,
  title={The regression analysis of binary sequences},
  author={Cox, David R},
  journal={Journal of the Royal Statistical Society: Series B},
  volume={20},
  number={2},
  pages={215--232},
  year={1958}
}

@article{Richman2000,
  author  = {Richman, Joshua S. and Moorman, J. Randall},
  title   = {Physiological time-series analysis using approximate entropy and sample entropy},
  journal = {American Journal of Physiology - Heart and Circulatory Physiology},
  volume  = {278},
  number  = {6},
  pages   = {H2039--H2049},
  year    = {2000},
  doi     = {10.1152/ajpheart.2000.278.6.H2039}
}

@article{Bandt2002,
  author  = {Bandt, Christoph and Pompe, Bernd},
  title   = {Permutation entropy: A natural complexity measure for time series},
  journal = {Physical Review Letters},
  volume  = {88},
  number  = {17},
  pages   = {174102},
  year    = {2002},
  doi     = {10.1103/PhysRevLett.88.174102}
}

@article{Higuchi1988,
  author  = {Higuchi, Tomoyuki},
  title   = {Approach to an irregular time series on the basis of the fractal theory},
  journal = {Physica D: Nonlinear Phenomena},
  volume  = {31},
  number  = {2},
  pages   = {277--283},
  year    = {1988},
  doi     = {10.1016/0167-2789(88)90081-4}
}

@article{Tzanetakis2002,
  author  = {Tzanetakis, George and Cook, Perry},
  title   = {Musical genre classification of audio signals},
  journal = {IEEE Transactions on Speech and Audio Processing},
  volume  = {10},
  number  = {5},
  pages   = {293--302},
  year    = {2002},
  doi     = {10.1109/TSA.2002.800560}
}

@book{Randall2011,
  author    = {Randall, Robert B.},
  title     = {Vibration-Based Condition Monitoring},
  publisher = {Wiley},
  year      = {2011},
  address   = {Chichester, UK},
  isbn      = {978-0470747858}
}

@article{Antoni2006,
  author  = {Antoni, Jerome},
  title   = {The spectral kurtosis: A useful tool for characterising non-stationary signals},
  journal = {Mechanical Systems and Signal Processing},
  volume  = {20},
  number  = {2},
  pages   = {282--307},
  year    = {2006},
  doi     = {10.1016/j.ymssp.2004.09.001}
}

@book{Devroye1996,
  author    = {Devroye, Luc and Gy{\"o}rfi, L{\'a}szl{\'o} and Lugosi, G{\'a}bor},
  title     = {A Probabilistic Theory of Pattern Recognition},
  publisher = {Springer},
  year      = {1996}
}

@book{Fukunaga1990,
  author    = {Fukunaga, Keinosuke},
  title     = {Introduction to Statistical Pattern Recognition},
  edition   = {2},
  publisher = {Academic Press},
  year      = {1990}
}

@article{Abellan2021,
  author  = {Abell{\'a}n, Joaqu{\'i}n and Castellano, Javier G. and R{\'i}os, Miguel},
  title   = {Analysis of the impact of class overlap on classification performance},
  journal = {Pattern Recognition},
  volume  = {117},
  pages   = {107999},
  year    = {2021},
  doi     = {10.1016/j.patcog.2021.107999}
}

@article{he2024indepth,
  title={In-Depth Insights into the Application of Recurrent Neural Networks (RNNs) in Traffic Prediction: A Comprehensive Review},
  author={He, Yuchen and Huang, Pin and Hong, Weihang and Luo, Qin and Li, Lishuai and Tsui, Kwok-Leung},
  journal={Algorithms},
  volume={17},
  number={9},
  pages={398},
  year={2024},
  publisher={MDPI},
  doi={10.3390/a17090398}
}

@article{fang2024dam,
  title={A Dam Displacement Prediction Method Based on a Model Combining Random Forest, a Convolutional Neural Network, and a Residual Attention Informer},
  author={Fang, Chunhui and Jiao, Ying and Wang, Xue and Lu, Taiqi and Gu, Hao},
  journal={Water},
  volume={16},
  number={24},
  pages={3687},
  year={2024},
  publisher={MDPI},
  doi={10.3390/w16243687}
}

@article{ye2024light,
  title={Light Recurrent Unit: Towards an Interpretable Recurrent Neural Network for Modeling Long-Range Dependency},
  author={Ye, Hong and Zhang, Yibing and Liu, Huizhou and Li, Xuannong and Chang, Jiaming and Zheng, Hui},
  journal={Electronics},
  volume={13},
  number={16},
  pages={3204},
  year={2024},
  publisher={MDPI},
  doi={10.3390/electronics13163204}
}

@inproceedings{hermiz2021impact,
  title={The Impact of Reducing Signal Acquisition Specifications on Neuronal Spike Sorting},
  author={Hermiz, John and Joseph, Erin and Lee, Kye Hyun and Baldacci, Ione A. and Chung, Jason E. and Frank, Loren M. and Bouchard, Kristofer E. and Denes, Peter},
  booktitle={2021 43rd Annual International Conference of the IEEE Engineering in Medicine \& Biology Society (EMBC)},
  pages={5914--5918},
  year={2021},
  organization={IEEE},
  doi={10.1109/EMBC46164.2021.9630669}
}

@article{bao2022effect,
  title={The Effect of Signal Duration on the Classification of Heart Sounds: A Deep Learning Approach},
  author={Bao, Xuan and Xu, Yanjun and Kamavuako, Ernest Nlandu},
  journal={Sensors},
  volume={22},
  number={6},
  pages={2261},
  year={2022},
  publisher={MDPI},
  doi={10.3390/s22062261}
}

\clearpage
\section*{Appendix}

\vspace{6pt}

\begin{center}

\small \textbf{Figure 7.} Dominant Amplitude ($z_1$)
\includegraphics[width=\columnwidth]{  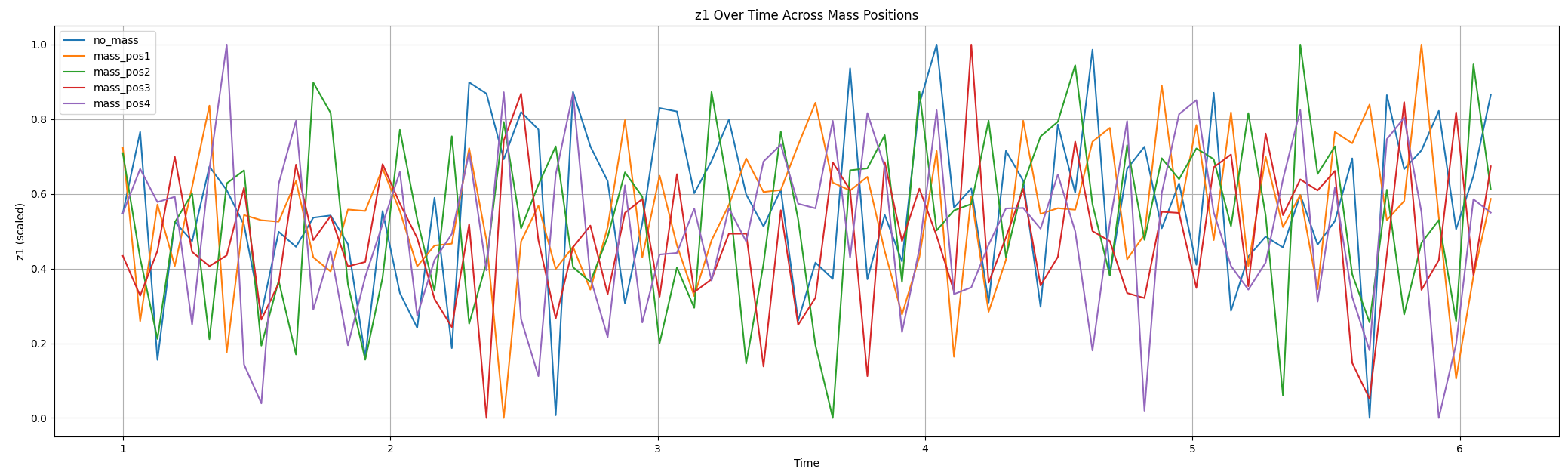}\\[-2mm]
\end{center}
\vspace{6pt}

\begin{center}
\small \textbf{Figure 8.} Sideband Symmetry ($z_2$)
\includegraphics[width=\columnwidth]{  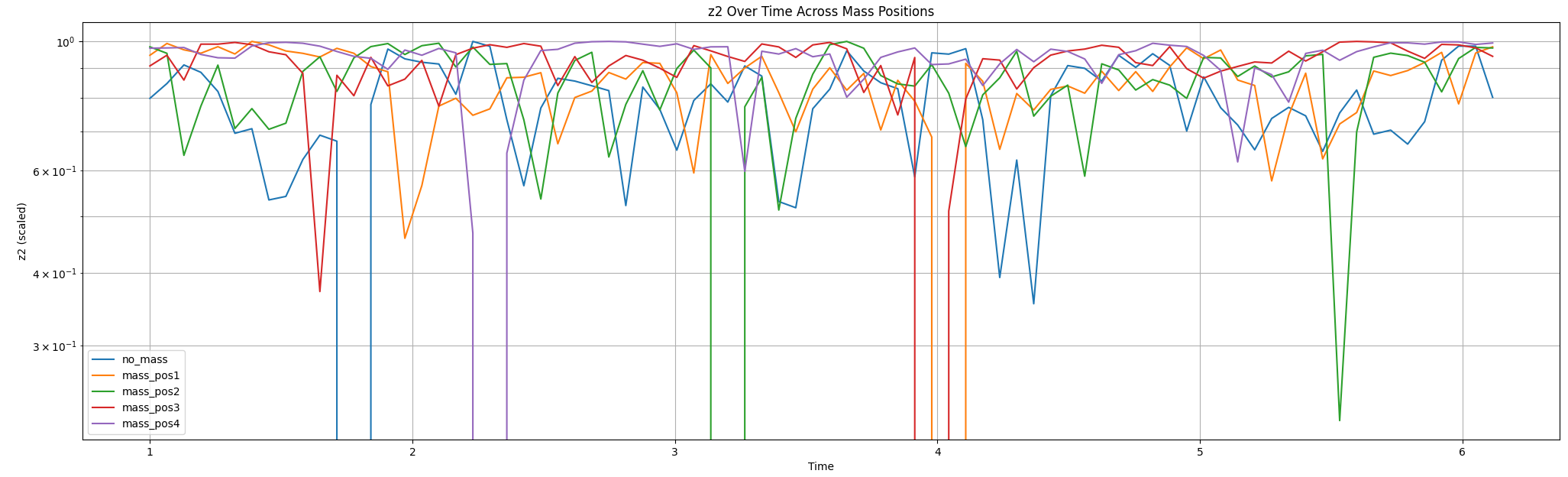}\\[-2mm]
\end{center}
\vspace{6pt}

\begin{center}
\small \textbf{Figure 9.} Second-peak Offset ($z_3$)
\includegraphics[width=\columnwidth]{  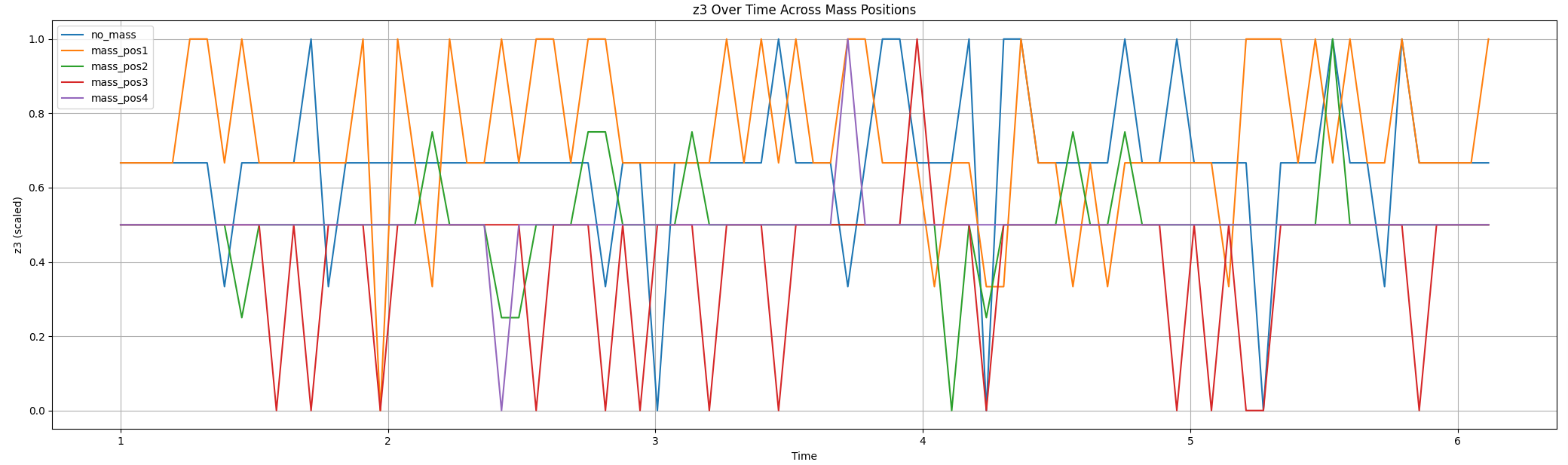}\\[-2mm]
\end{center}
\vspace{6pt}

\begin{center}
\small \textbf{Figure 10.} Harmonic Ratio ($z_4$)
\includegraphics[width=\columnwidth]{  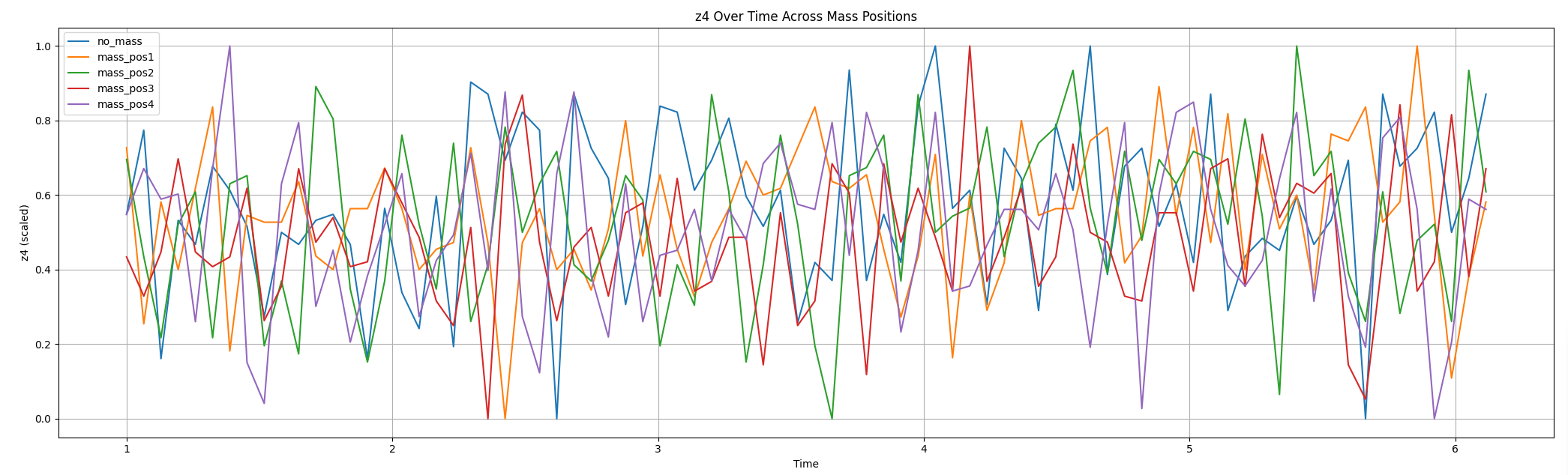}\\[-2mm]
\end{center}
\vspace{6pt}

\begin{center}
\small \textbf{Figure 11.} Continuous Wavelet 
\includegraphics[width=\columnwidth]{  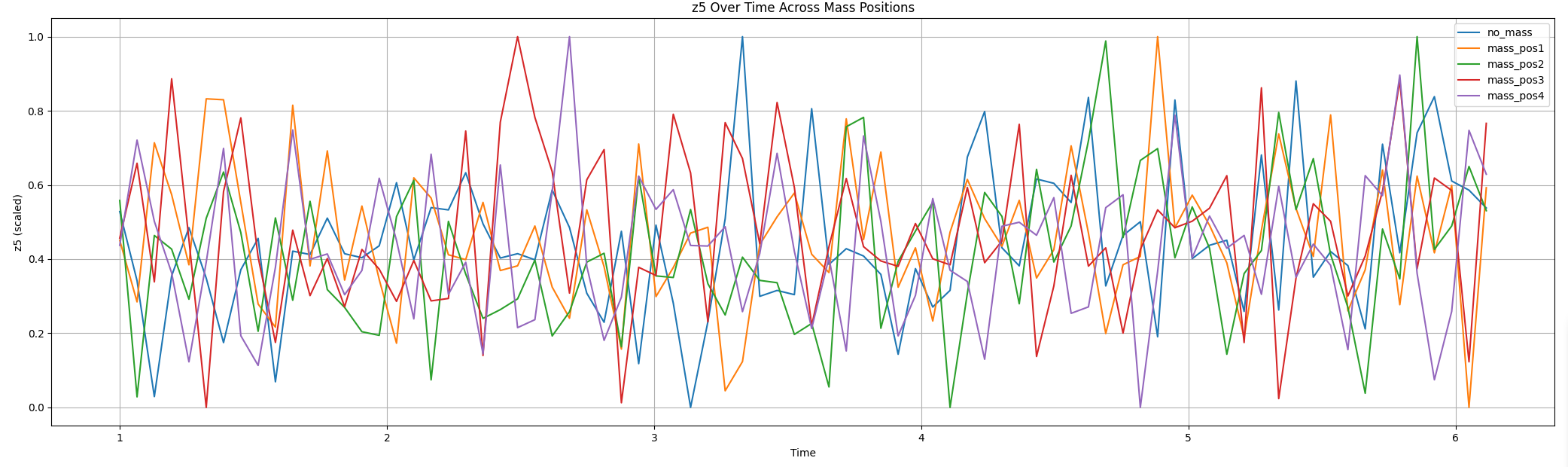}\\[-2mm]
Transform ($z_5$)
\end{center}
\vspace{6pt}

 \begin{center}
 \small \textbf{Figure 12.} CEEMDAN Energy Ratio ($z_6$)
\includegraphics[width=\columnwidth]{  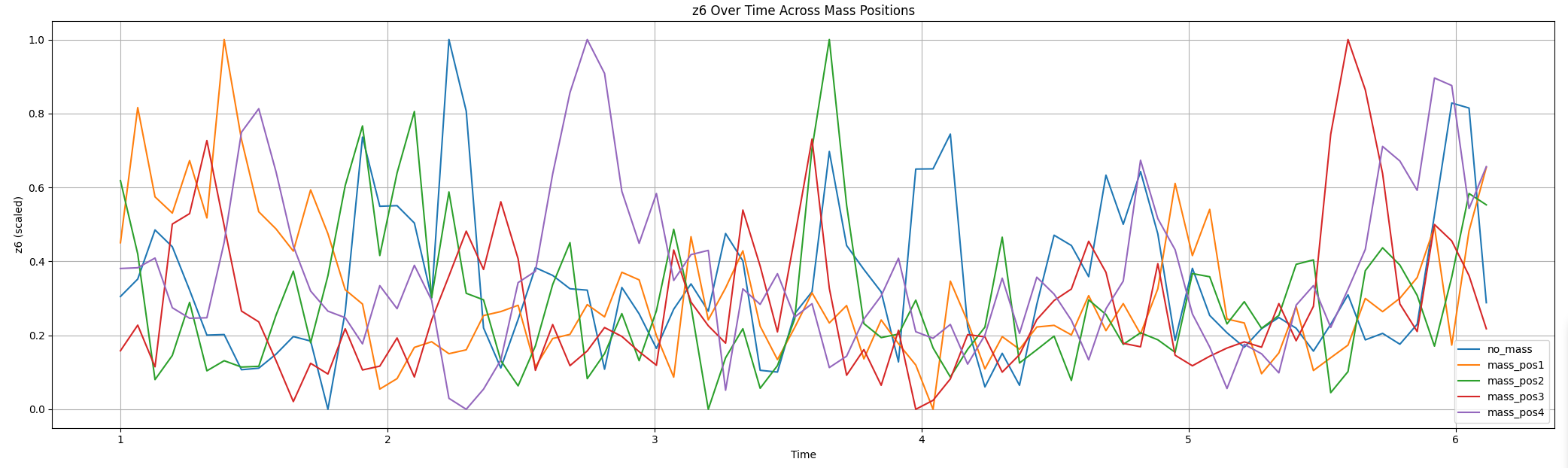}\\[-2mm]

\end{center}

\refstepcounter{figure}\label{fig:signals}

\noindent
\small \textbf{Figure 13.} FFT amplitude spectra of the five vibration signals. 
The spectra indicate that $\delta_{3}$ and $\delta_{4}$ are relatively well separated, 
whereas $\eta$, $\delta_{1}$, and $\delta_{2}$ exhibit substantial overlap, making them difficult to distinguish. 
Such spectral overlap can reduce classification accuracy in multi-class machine-learning-based signal analysis.

\begin{center}
\includegraphics[width=\columnwidth]{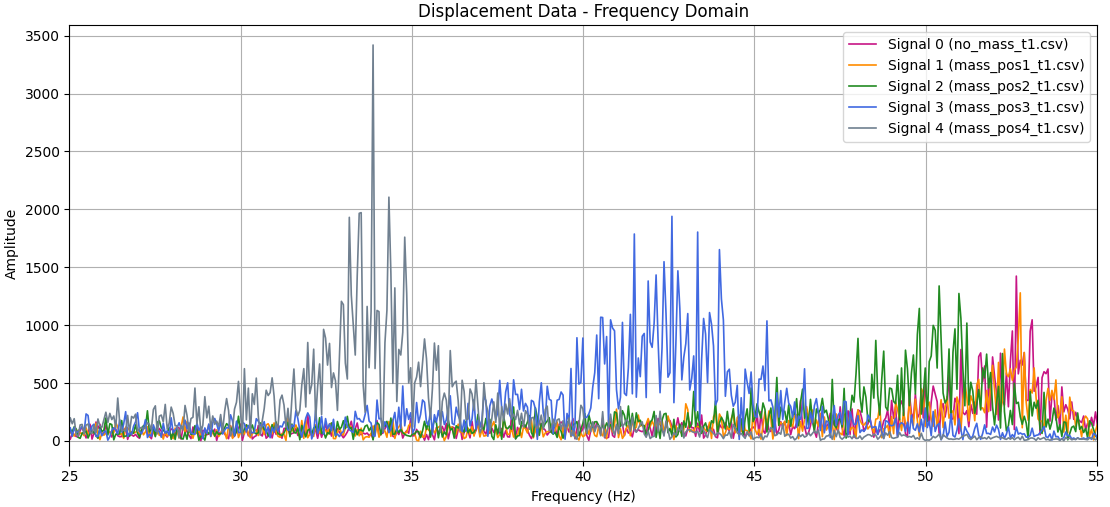}
\end{center}

\refstepcounter{figure}\label{fig:rawsignals}`

\noindent
\small \textbf{Figure 14.} Original signals in the time--displacement domain. 
Although each state (healthy, damaged-1 to damaged-4) is shown, the signals appear highly similar, 
making it difficult to distinguish conditions without further analysis.

\vspace{3pt}

\begin{center}
\includegraphics[width=\columnwidth]{  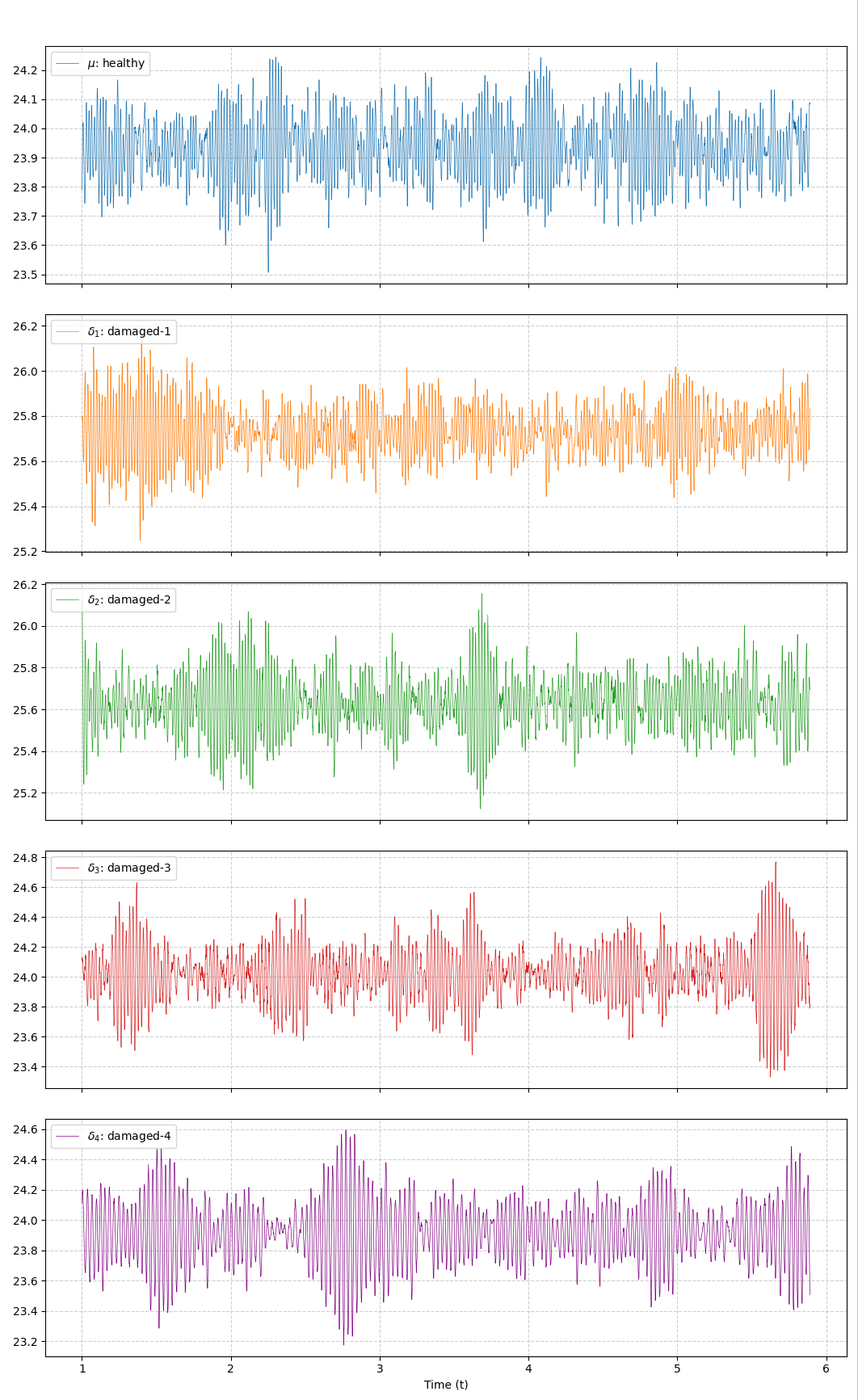}
\end{center}

\end{document}